\pdfoutput=1
\documentclass[11pt]{article}
\usepackage[final]{acl}

\usepackage{times}
\usepackage{latexsym}

\usepackage[T1]{fontenc}

\usepackage[utf8]{inputenc}
\usepackage{float}
\usepackage{microtype}
\usepackage{amssymb}
\usepackage{lineno}
\usepackage{tikz}
\usetikzlibrary{arrows.meta}
\usepackage{array}
\usepackage{times}
\usepackage{helvet}
\usepackage{courier}
\usepackage{multirow}
\usepackage{mathrsfs}
\usepackage{graphicx}
\usepackage{enumitem}
\usepackage{graphicx}
\usepackage{blindtext}
\usepackage{algorithm}
\usepackage{algorithmic}
\usepackage[marginal]{footmisc}
\usepackage[utf8]{inputenc}
\usepackage[english]{babel}
\usepackage{epstopdf}
\usepackage{array}
\usepackage{multirow}
\usepackage{booktabs}
\usepackage{amsmath}
\usepackage{wrapfig}
\usepackage{inconsolata}
\usepackage{url}

\usepackage{url}
\usepackage{multirow}
\usepackage{svg}

\usepackage{xcolor}

\definecolor{lightpink}{HTML}{ed9782}
\definecolor{lightblue}{HTML}{5395f5}
\definecolor{lightgreen}{HTML}{efd08f}
\definecolor{grey}{HTML}{b3b3b3}
\definecolor{cgreen}{RGB}{46, 139, 87} 

\definecolor{lightorange}{RGB}{255, 190, 123}
\usepackage[most]{tcolorbox}

\usepackage{algorithm}
\usepackage{algorithmic}

\usepackage{enumitem}
\usepackage{tabularx}
\usepackage{subcaption}
\usepackage{makecell}

\usepackage{tabularx}
\usepackage{multirow}
\usepackage[normalem]{ulem}
\useunder{\uline}{\ul}{}

\usepackage{times}
\usepackage{moreverb}
\usepackage{graphicx}
\usepackage{mathrsfs}
\usepackage{bm}
\usepackage{url}
\usepackage{amsmath}
\usepackage{graphicx}

\title{Does Deeper Reasoning Compromise Alignment? Revealing and Mitigating of Alignment Collapse in Large Reasoning Models}

\author{
  Yu-Hang Wu\textsuperscript{1},
  Yu-Jie Xiong\textsuperscript{1}\thanks{ Corresponding author.},
  Henghua Zhang\textsuperscript{1},
  Bairui Zhang\textsuperscript{2},
  Jia-Chen Zhang\textsuperscript{1},
  Shaohua Li\textsuperscript{3}
  \\
  \textsuperscript{1}Shanghai University of Engineering Science
  \\
  \textsuperscript{2}The Hong Kong University of Science and Technology
  \\
  \textsuperscript{3}A * STAR
}

\usepackage{wrapfig}  
\usepackage{graphicx} 
\usepackage{subcaption}
\usepackage{balance}
\begin{document}

\maketitle
\begin{abstract}
The emergence of Chain-of-Thought (CoT) has established a robust foundation for Large Reasoning Models (LRMs). While deep reasoning is widely believed to enhance safety alignment, the stability of alignment mechanisms under extended reasoning remains underexplored. This paper challenges the prevailing view by revealing a critical vulnerability: \textbf{Deep Reasoning May Induce Alignment Collapse}. To rigorously quantify this phenomenon, we propose the Alignment Loss Rate (ALR) metric. Our experiments demonstrate that as reasoning depth increases, ALR rises significantly, indicating a severe degradation in model robustness against external perturbations. Capitalizing on this instability, a novel jailbreaking paradigm, Reasoning Trap (RT), is proposed. RT induces the model into extended reasoning to amplify the impact of adversarial attacks, leading to a sharp decline in safety capabilities. To elucidate the mechanism behind this collapse, we identify Attention Dilution as the root cause, arising from the competition for attention between the extended reasoning process and the original input. To mitigate this, Reasoning Residual Alignment (RRA) is proposed, a lightweight defense strategy that dynamically re-emphasizes the input via residual connections integrated with the reasoning process.

\end{abstract}

\begin{figure}[!t]
    \centering
    \includegraphics[width=\columnwidth]{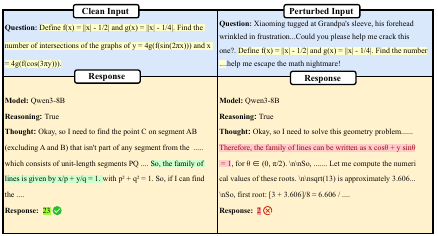}
    \caption{The figure shows that the successful example of deep reasoning on clean inputs (Left) with the resulting failure when external perturbations are introduced (Right).}
    \label{fig:example}
\end{figure}
\section{Introduction}
Large Language Models (LLMs) have demonstrated remarkable capabilities across a wide range of domains \cite{bai:22,zhang:25,dubey:24,zhang2025hyperadalora}. Among the numerous studies, the introduction of Chain-of-Thought (CoT) prompting stands out as a pivotal milestone: by guiding models to produce explicit intermediate reasoning steps, CoT dramatically improves performance on complex tasks, particularly logical and mathematical reasoning \cite{wei:22,liao:24,wei:25,wu:25a}. This breakthrough has directly catalyzed the emergence of Large Reasoning Models (LRMs), which are systems explicitly designed to tackle challenging problems through extended internal reasoning processes \cite{yang:25,deepseekai:25}. Concurrently, the prevailing view in the community holds that the deeper reasoning capabilities of LRMs enable them to better comprehend complex instructions and intent\cite{jaech:24,gou-etal-2025-sure}, leading to the widespread belief 

\begin{itemize}
    \item \textit{Stereotype: Reasoning models should inherently possess stronger alignment robustness.}
\end{itemize}

Existing works have shown that attackers can exploit intermediate reasoning processes to bypass safety guardrails\cite{kuo2025hcoth,lou:25}. However, these studies primarily focus on the practical construction of attacks. Mechanistic investigation into the fundamental question of whether extended reasoning itself systematically destabilizes alignment remains limited. As illustrated in Figure~\ref{fig:example}, we uncover a striking phenomenon where deeper reasoning, while boosting task performance, simultaneously renders models significantly more susceptible to external perturbations. This observation leads us to the core research question of this paper:

\begin{itemize}
    \item \textit{RQ: Does the intrinsic instability induced by extended deep reasoning compromise the alignment robustness of LRMs?}
\end{itemize}

To answer this question, we first conduct a systematic investigation into the alignment robustness  of LRMs. We reveal a critical vulnerability termed Alignment Collapse: 

\begin{itemize}
    \item \textit{Finding: While deep reasoning enhances performance on standard tasks, it significantly degrades the model's ability to adhere to original constraints under external perturbations.}
\end{itemize}

Specifically, our evaluation using a proposed Alignment Loss Rate (ALR) metric demonstrates that as reasoning depth increases, models exhibit a systematic deviation from their intended alignment, rendering them increasingly fragile.

Building on this finding, we extend our investigation to the safety domain. We propose RT (Reasoning Trap) attack to validate that the identified collapse creates exploitable risks. Results show that by inducing extended reasoning, RT amplifies existing adversarial attacks, precipitating a sharp decline in safety capabilities. Mechanistically, we attribute this intrinsic instability to Attention Dilution. Our analysis uncovers that during extended autoregressive generation, the reasoning process structurally competes for attention weight against the original input. This competition causes attention weights on the original input to decay rapidly, rendering the final generation phase highly susceptible to external perturbations, thereby compromising alignment robustness. Finally, to mitigate this, we propose RRA (Reasoning Residual Alignment). Distinguished by being training-free, RRA dynamically re-emphasizes input via residual connections to effectively enhance alignment robustness. The main contributions of our paper can
be summarized as follows:

\begin{itemize} 
 \item \textbf{Revelation of Alignment Collapse:} We are the first to identify a critical phenomenon in LRMs: extended reasoning significantly degrades the model's adherence to alignment under perturbation. We propose the ALR to systematically quantify this robustness decay.
\item \textbf{Interpretable Analysis:}  We attribute this collapse to Attention Dilution. Our analysis reveals that the structural competition for attention resources during long-context reasoning leads to the rapid decay of attention weights on the original input.
\item \textbf{Risk and Strategy:} We demonstrate that this instability significantly compromises safety alignment via RT, which exploits extended reasoning to amplify attacks, and propose RRA, a training-free defense requiring no additional prompting that effectively restores alignment robustness.
\end{itemize}

\section{Related Work}
\subsection{Exploration of LRMs}
The pursuit of superior reasoning capabilities has emerged as a central frontier in LLMs, epitomized by the advent of models such as DeepSeek \cite{deepseekai:25} and Qwen3 \cite{yang:25}. Research in this domain generally bifurcates into two paradigms: inference-time strategies, such as CoT prompting \cite{wei:22} and Tree of Thoughts \cite{yao:23}, which serve as scaffolding techniques to elicit latent multi-step reasoning abilities without altering model weights; and training-time methodologies, which focus on internalizing reasoning processes directly into model parameters via Supervised Fine-Tuning (SFT) \cite{zhang:25a} or Reinforcement Learning (RL) \cite{yao:23,chen:25}, enabling models to natively generate extended thought processes for solving complex tasks. While deep reasoning has significantly enhanced performance in logical and mathematical domains, its impact on safety alignment remains a subject of intense debate. Recent studies challenge the assumption that reasoning inherently improves safety: \cite{kuo2025hcoth} demonstrate that attackers can exploit reasoning processes to introduce new vulnerabilities, and \cite{lou:25} observe safety degradation in Multimodal Large Reasoning Models (MLRMs). However, existing literature is largely confined to phenomenological revelations or the verification of specific attacks. This lack of systematic analysis makes it critically important to mechanistically clarify how deep reasoning compromises alignment stability to ensure the development of robust and secure reasoning models.

\begin{table*}[t]
\centering
\renewcommand{\arraystretch}{1.2}
\setlength{\tabcolsep}{14pt}
\scalebox{0.9}{%
\begin{tabular}{@{}lccc@{}}
\toprule
\textbf{Setting ($L$)} & \textbf{Clean} & \textbf{Perturbation-I (Noise)} & \textbf{Perturbation-II (Nesting)} \\
\midrule
Qwen3-8B ($L=0$, no-reasoning)   & 20.83 & 18.75 \textbf{(\textcolor{lightgreen}{-02.08})} & 14.16 \textbf{(\textcolor{lightgreen}{-06.67})} \\
Qwen3-8B ($L=2048$, Reasoning)   & 24.58 & 20.41 \textbf{(\textcolor{lightgreen}{-04.17})} & 15.42 \textbf{(\textcolor{lightgreen}{-09.16})} \\
Qwen3-8B ($L=4096$, Reasoning)   & 31.67 & 25.00 \textbf{(\textcolor{lightgreen}{-06.67})} & 19.16 \textbf{(\textcolor{lightgreen}{-12.51})} \\
Qwen3-14B ($L=0$, no-reasoning)   & 31.66 & 29.58 \textbf{(\textcolor{lightgreen}{-02.08})} & 28.75 \textbf{(\textcolor{lightgreen}{-02.91})} \\
Qwen3-14B ($L=2048$, Reasoning)   & 40.00 & 35.41 \textbf{(\textcolor{lightgreen}{-04.59})} & 31.66 \textbf{(\textcolor{lightgreen}{-08.34})} \\
Qwen3-14B ($L=4096$, Reasoning)   & 49.16 & 43.33 \textbf{(\textcolor{lightgreen}{-05.83})} & 32.91 \textbf{(\textcolor{lightgreen}{-16.25})} \\
\bottomrule
\end{tabular}%
}

\caption{Task accuracy (\%) under different reasoning depths ($L$) on AIME2024 dataset\cite{MAA:24}. \textbf{Clean} denotes standard performance $\mathcal{A}(L)$ on original inputs, while the perturbation columns represent $\mathcal{A}'(L)$. \textbf{Perturbation-I} introduces irrelevant nonsense text (Noise), while \textbf{Perturbation-II} wraps instructions within nested scenarios (Nesting). Values in \textbf{\textcolor{lightgreen}{lightgreen}} parentheses indicate the absolute accuracy drop ($\mathcal{A}'(L) - \mathcal{A}(L)$) relative to the Clean baseline. Detailed descriptions of Perturbation-I and Perturbation-II can be found in Appendix\ref{sec:exp_set}.}
\label{tab:alignment_finding}
\end{table*}

\subsection{Jailbreak Attacks on LLMs}
Existing jailbreaking attacks have been extensively applied to LLMs. Early manual techniques, such as DAN~\cite{Shen:23}, demonstrated the effectiveness of role-playing prompts in evading safeguards. Subsequent research systematized these methods by classifying them according to tactics, objectives, and capability-safety balances~\cite{Wu2025SugarCoatedPB}. Gradient-based optimization approaches, including GCG~\cite{Zou2023UniversalAT}, AutoDAN~\cite{liu:24}, and I-GCG~\cite{jia:25}, iteratively craft adversarial suffixes but incur high computational costs. In contrast, heuristic methods offer greater efficiency at the expense of consistency~\cite{kuo2025hcoth}, while LLM-assisted frameworks like PAIR~\cite{chao:23}, FlipAttack~\cite{liu:25}, and PAP~\cite{zeng:24} leverage auxiliary models to streamline prompt refinement and enhance scalability. Despite these advances, universal jailbreaks remain challenging amid evolving defensive strategies~\cite{zhu:25}. Therefore, adapting these attacks to exploit the alignment collapse induced by deep reasoning in LRMs, presents a promising avenue for improving both efficiency and attack success rates.

\section{The Alignment Issues in LRM}
\label{sec:issue}
This section aims to systematically investigate the potential impact of deep reasoning on the alignment robustness of LRMs. We define the key notations and experimental setup. Subsequently, through controlled experiments introducing external perturbations into reasoning tasks, we reveal a phenomenon: While deep reasoning yields performance gains, it may simultaneously induce a significant degradation in the model's alignment robustness.

\begin{figure*}[!t] 
    \centering
    \subfloat[Qwen3-8B (AIME2024)\label{fig:alignment_loss_8b}]{
        \includegraphics[width=0.42\textwidth]{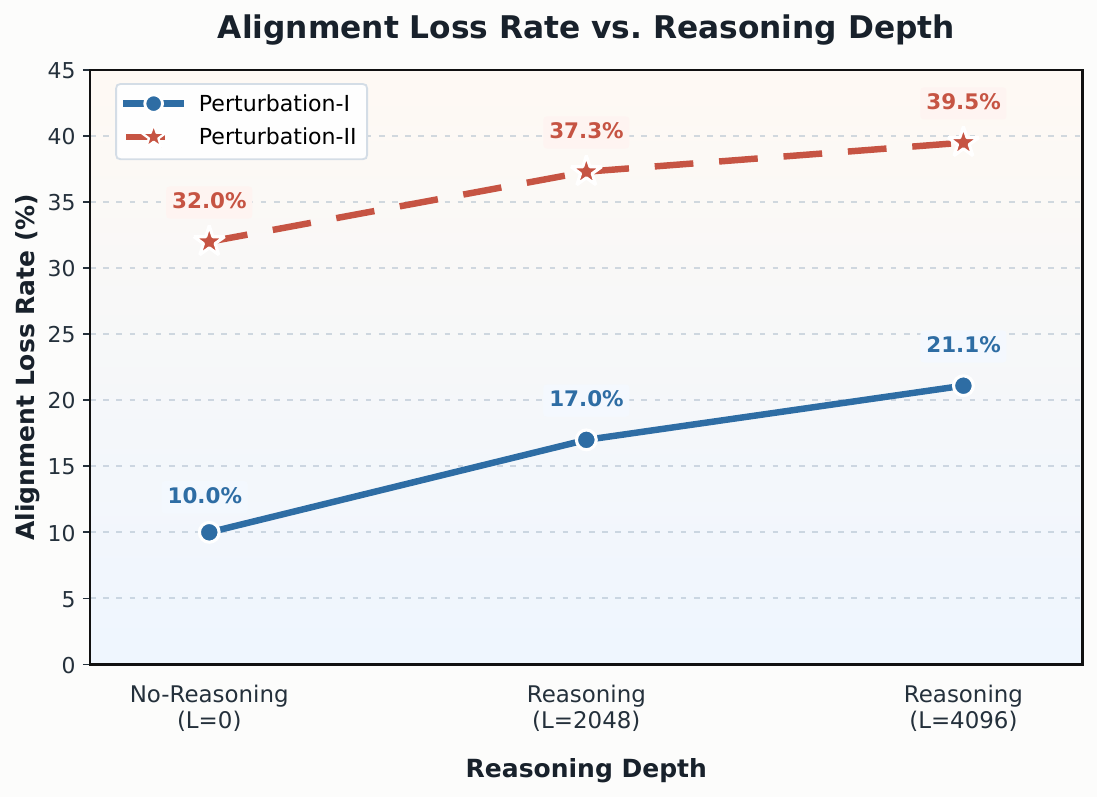} 
    }
    \hspace{0.04\textwidth} 
    \subfloat[Qwen3-14B (AIME2024)\label{fig:alignment_loss_14b}]{
        \includegraphics[width=0.42\textwidth]{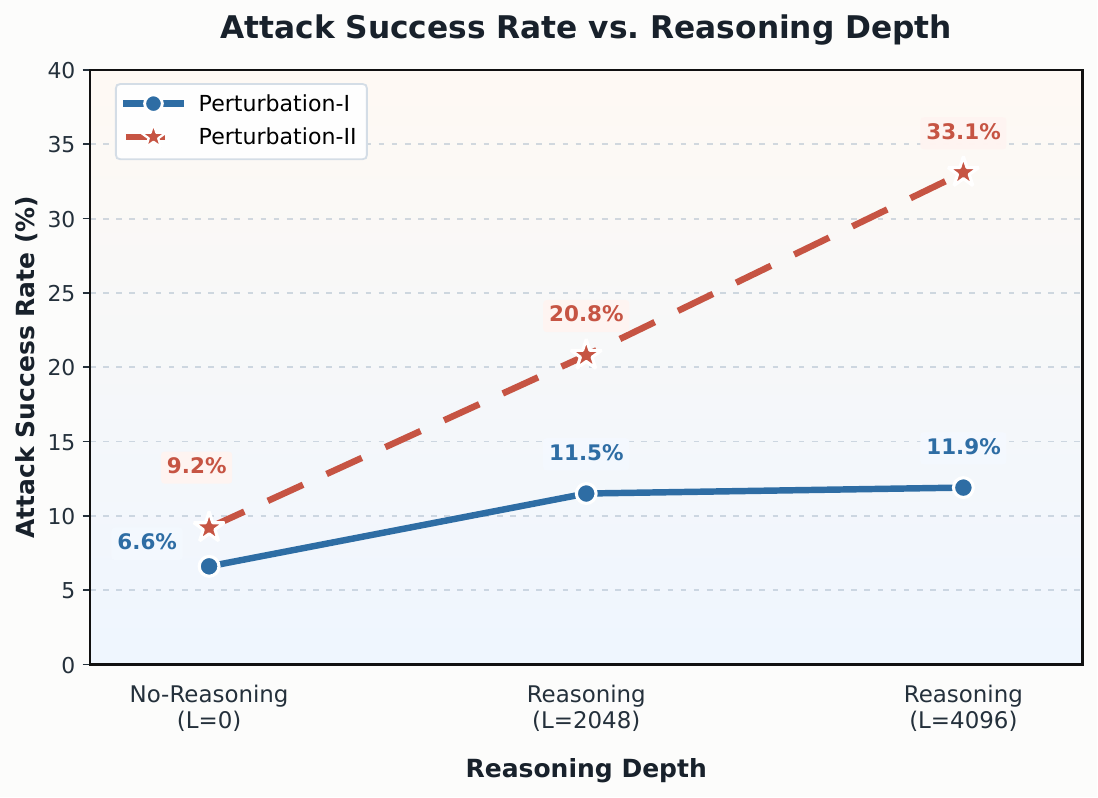}
    }
    \caption{The trend of \textbf{Alignment Loss Rate (ALR)} across varying reasoning depths ($L$) for (a) Qwen3-8B and (b) Qwen3-14B on AIME2024 dataset. The monotonic increasing trend indicates that deep reasoning amplifies the relative alignment degradation under perturbations.}
    \label{fig:alignment_loss_comparison} 
\end{figure*}
\begin{table}[t] 
\centering
\renewcommand{\arraystretch}{1.2}  
\setlength{\tabcolsep}{12pt}        
\scalebox{0.8}{%
\begin{tabular}{@{}lcc@{}}
\toprule
\textbf{Setting($L$)} & \textbf{Clean} & \textbf{Perturbation-I} \\
\midrule
Claude-Haiku-4.5 ($L$ = 0)    & 53.33             & 51.66\textbf{(\textcolor{lightgreen}{-1.67})}             \\
Claude-Haiku-4.5 ($L$ = 2048)  & 57.50              & 55.00\textbf{(\textcolor{lightgreen}{-2.50})}             \\
Claude-Haiku-4.5 ($L$ = 4096)  & 61.66  & 58.33\textbf{(\textcolor{lightgreen}{-3.33})} \\
\bottomrule
\end{tabular}%
}
\caption{Task accuracy (\%) on AIME dataset evaluated with Claude-Haiku-4.5 across varying reasoning depth $L$. Parentheses in \textbf{\textcolor{lightgreen}{lightgreen}} indicate the absolute performance drop under Perturbation-I relative to the Clean baseline.}
\label{tab:add_exp}
\end{table}
\subsection{Preliminary}
\noindent \textbf{Models and Datasets.} We employ Qwen3 as the primary experimental model, because it is currently the unique open-source model capable of flexibly switching between non-reasoning and deep reasoning modes of varying depths. This capability provides an ideal controlled environment for comparative analysis. Our experiments are conducted on the AIME2024\cite{MAA:24} and LogicAsker\cite{wan:24} datasets, which are specifically designed to evaluate the reasoning capabilities of LLMs.

\noindent \textbf{Notation Define.} we denote the target model as $\mathcal{M}_t$ and the sequence length limit for deep reasoning as $L$. Specifically, $L=0$ is defined as the non-reasoning mode, while $L=2048$ and $L=4096$ represent reasoning modes of varying depths. Given an evaluation dataset $\mathcal{D} = \{(x_i, y_i)\}_{i=1}^N$, where $x_i$ is the clean input and $y_i$ is the ground truth. To rigorously quantify the impact of deep reasoning on alignment robustness, we define the model's inference process under two conditions: the standard inference on clean inputs and the perturbed inference under the external perturbation function $\mathcal{F}(\cdot)$. The generated outputs for the $i$-th sample are formulated as:

\[
\begin{aligned}
\tilde{y}_i &= \mathcal{M}_t(x_i; L), \\
\tilde{y}'_i &= \mathcal{M}_t(\mathcal{F}(x_i); L),
\end{aligned} \tag{1}
\]
where $\tilde{y}_i$ and $\tilde{y}'_i$ denote the responses generated from the original and perturbed inputs, respectively. Subsequently, the correctness of the generated responses is determined via hard matching with $y_i$, resulting in a binary indicator $\mathcal{C}_{\text{eval}}(\cdot, y_i) \in \{True, False\}$.

The task accuracy under the reasoning depth $L$ is calculated by aggregating the evaluation scores over the entire dataset. We define $\mathcal{A}(L)$ as the model's performance on clean inputs and $\mathcal{A}'(L)$ as the performance on perturbative inputs:

\[
\begin{aligned}
\mathcal{A}(L) &= \frac{1}{N} \sum_{i=1}^{N} \mathcal{C}_{\text{eval}}(\tilde{y}_i, y_i), \\
\mathcal{A}'(L) &= \frac{1}{N} \sum_{i=1}^{N} \mathcal{C}_{\text{eval}}(\tilde{y}_i, y_i). \\
\end{aligned} \tag{2}
\]

We define the Relative Alignment Loss Rate (${\text{ALR}}(L)$) as the percentage degradation in accuracy relative to the model's original capability:

\[
\begin{aligned}
\text{ALR}(L) &= \frac{\mathcal{A}(L) - \mathcal{A}'(L)}{\mathcal{A}(L)} \times 100\%.
\end{aligned} \tag{3}
\label{eq:delta_loss}
\]

Consequently, $\text{ALR}(L)$ serves as our primary metric to quantify Alignment Collapse. A higher value indicates that external deep reasoning causes a larger proportional deviation from the model's original capabilities, signifying compromised alignment robustness.

\begin{table}[t] 
\centering
\renewcommand{\arraystretch}{1.2}  
\setlength{\tabcolsep}{8pt}        
\scalebox{0.8}{%
\begin{tabular}{@{}lcc@{}}
\toprule
\textbf{Setting($L$)} & \textbf{Clean} & \textbf{Perturbation-I (Noise)} \\
\midrule
Qwen3-8B ($L$ = 0)    & 57.94              & 53.71\textbf{(\textcolor{lightgreen}{-04.23})}             \\
Qwen3-8B ($L$ = 2048)  & 65.48              & 55.21\textbf{(\textcolor{lightgreen}{-10.27})}             \\
Qwen3-8B ($L$ = 4096)  & 69.77  & 58.12\textbf{(\textcolor{lightgreen}{-11.65})} \\
\midrule  
Qwen3-14B ($L$ = 0)   & 56.55  & 55.22\textbf{(\textcolor{lightgreen}{-01.33})} \\
Qwen3-14B ($L$ = 2048) & 82.84  & 79.63\textbf{(\textcolor{lightgreen}{-03.21})}             \\
Qwen3-14B ($L$ = 4096) & 83.68              & 73.28\textbf{(\textcolor{lightgreen}{-10.40})}             \\
\midrule
GLM4.6 ($L$ = 0)    & 98.20            & 98.10\textbf{(\textcolor{lightgreen}{-0.10})}             \\
GLM4.6 ($L$ = 2048)  & 99.65              & 99.38\textbf{(\textcolor{lightgreen}{-0.27})}             \\
GLM4.6 ($L$ = 4096)  & 100.00  & 99.70\textbf{(\textcolor{lightgreen}{-0.30})} \\
\bottomrule
\end{tabular}%
}
\caption{Task accuracy (\%) on LogicAsker\cite{wan:24} across varying $L$. Parentheses in \textbf{\textcolor{lightgreen}{lightgreen}} indicate the absolute performance drop under Perturbation-I (Noise) relative to the Clean baseline. The ALR trend can be found in Appendix~\ref{sec:add_exp}.}
\label{tab:method_comparison}
\end{table}

\begin{table*}[t]
\centering
\renewcommand{\arraystretch}{1.2}
\setlength{\tabcolsep}{8pt}
\scalebox{0.84}{%
\begin{tabular}{@{}lccccccc@{}}
\toprule
\multirow{2}{*}{\textbf{Method}} & 
\multicolumn{3}{c}{\textbf{Reasoning Mode}} & 
\multicolumn{3}{c}{\textbf{Standard Mode}} & 
\multirow{2}{*}{\textbf{Average}} \\ 
\cmidrule(lr){2-4} \cmidrule(lr){5-7}
 & \textbf{DeepSeek-R1} & \textbf{Qwen3-8B} & \textbf{GLM4.6} & 
   \textbf{DeepSeek-V3} & \textbf{Qwen3-8B} & \textbf{GLM4.6} & \\ 
\midrule
No Attack    
& 99.62 \textbf{(\textcolor{lightpink}{-00.38})} 
& 99.81 \textbf{(\textcolor{lightpink}{-00.19})} 
& 99.73 \textbf{(\textcolor{lightpink}{-00.27})} 
& 100.00 & 100.00 & 100.00 & 00.28 \\

FlipAttack   
& 02.19 \textbf{(\textcolor{lightpink}{-01.07})} 
& 57.42 \textbf{(\textcolor{lightpink}{-34.89})} 
& 70.58 \textbf{(\textcolor{lightpink}{-24.61})} 
& 03.26 & 92.31 & 95.19 & 20.19 \\

PAP          
& 87.69 (+00.19) 
& 73.07 \textbf{(\textcolor{lightpink}{-12.50})} 
& 84.04 \textbf{(\textcolor{lightpink}{-06.34})} 
& 87.50 & 85.57 & 90.38 & 06.22 \\

ArtPrompt    
& 28.46 \textbf{(\textcolor{lightpink}{-07.50})} 
& 40.00 \textbf{(\textcolor{lightpink}{-06.35})} 
& 39.62 \textbf{(\textcolor{lightpink}{-18.46})} 
& 35.96 & 46.35 & 58.08 & 10.77 \\
\bottomrule
\end{tabular}%
}
\caption{Comparison of Rejection Success Rates (RSR, \%) between Reasoning ($L=4096$) and Standard ($L=0$) Modes on AdvBench~\cite{Zou2023UniversalAT}. Values in \textbf{\textcolor{lightpink}{light pink}} show the drop in RSR in Reasoning Mode compared to Standard Mode, indicating safety degradation under deep reasoning.}
\label{tab:reasoning_susceptibility_avg}
\end{table*}

\subsection{Deeper Reasoning Induces Alignment Collapse}
\noindent \textbf{The Performance and Alignment Robustness.} Table \ref{tab:alignment_finding} reveals a phenomenon where reasoning depth enhances standard capabilities but amplifies alignment vulnerabilities under perturbation. On clean inputs, Qwen3-8B achieves progressive gains, climbing from 20.83\% ($L=0$) to 24.58\% ($L=2048$) and 31.67\% ($L=4096$). However, this benefit is compromised by a depth-dependent increase in fragility. While the non-reasoning baseline ($L=0$) remains relatively resilient with a drop of 6.67\% (Perturbation-II), the degradation intensifies with depth: the absolute drop widens to 9.16\% at $L=2048$ and escalates to 12.51\% at $L=4096$. This stepwise deterioration indicates that the model's alignment robustness becomes increasingly susceptible to interference as reasoning extends. As shown in Table~\ref{tab:method_comparison}, we observe similar trends in the larger model and across the LogicAsker dataset.

\noindent \textbf{Quantifying Alignment Collapse.} To rigorously quantify the relationship between reasoning depth and robustness, Figure \ref{fig:alignment_loss_comparison} visualizes the ALR across varying depths. The trend is unmistakably monotonic as ALR exhibits a strong positive correlation with reasoning depth $L$. Taking Qwen3-14B as a primary example, the ALR starts at a negligible 9.2\% at $L=0$. As the reasoning depth increases to $L=2048$, the ALR climbs significantly to 20.8\%, and ultimately surges to 33.1\% at $L=4096$. This consistent escalation confirms that deeper reasoning structurally undermines the model's alignment robustness. This empirical evidence supports a finding: 

\begin{itemize}
    \item \textit{Deep reasoning enhances task performance but compromise alignment robustness.}
\end{itemize}
Under identical perturbation conditions, models with greater reasoning depth suffer increasingly severe performance degradation. We refer to this phenomenon as Alignment Collapse.


\section{Reasoning Trap: Uncovering Safety Risks in Deep Reasoning}
\label{sec:drj}
Motivated by the alignment collapse observed in Section~\ref{sec:issue}, 
we introduce Reasoning Trap (RT), a reasoning-triggered framework for revealing safety risks in standard LLMs. 
RT combines existing jailbreak prompts with a reasoning trigger to test whether extended reasoning amplifies adversarial perturbations and weakens refusal behavior.

\begin{table*}[t]
    \centering
    \renewcommand{\arraystretch}{1.2}
    \setlength{\tabcolsep}{8pt}
    \scalebox{0.9}{
        \begin{tabular}{@{}lccccc@{}}
            \toprule
            \textbf{Method} & 
            \textbf{Qwen3-8B} $\downarrow$ & 
            \textbf{Qwen3-14B} $\downarrow$ & 
            \textbf{Llama2-7B} $\downarrow$ & 
            \textbf{Llama2-13B} $\downarrow$ & 
            \textbf{DeepSeek-V3} $\downarrow$ \\
            \midrule
            No Attack+RT & 
            98.27\textbf{(\textcolor{lightpink}{-01.73})} & 
            99.42\textbf{(\textcolor{lightpink}{-00.58})} & 
            100.00\textbf{(\textcolor{lightpink}{-00.00})} & 
            100.00\textbf{(\textcolor{lightpink}{-00.00})} & 
            100.0\textbf{(\textcolor{lightpink}{-00.00})} \\
            
            FlipAttack+RT & 
            29.03\textbf{(\textcolor{lightpink}{-63.28})} & 
            45.57\textbf{(\textcolor{lightpink}{-02.12})} & 
            100.00\textbf{(\textcolor{lightpink}{-00.00})} & 
            100.00\textbf{(\textcolor{lightpink}{-00.00})} & 
            00.38\textbf{(\textcolor{lightpink}{-02.88})} \\
            
            PAP+RT & 
            81.53\textbf{(\textcolor{lightpink}{-04.04})} & 
            80.75\textbf{(\textcolor{lightpink}{-01.94})} & 
            88.46\textbf{(\textcolor{lightpink}{-06.35})} & 
            89.24\textbf{(\textcolor{lightpink}{-03.07})} & 
            79.42\textbf{(\textcolor{lightpink}{-08.08})} \\
            
            ArtPrompt+RT & 
            38.84\textbf{(\textcolor{lightpink}{-07.51})} & 
            35.00\textbf{(\textcolor{lightpink}{-26.54})} & 
            36.54\textbf{(\textcolor{lightpink}{-43.65})} & 
            37.21\textbf{(\textcolor{lightpink}{-26.06})} & 
            21.15\textbf{(\textcolor{lightpink}{-14.81})} \\
            \bottomrule
        \end{tabular}
    }
    \caption{Performance of RT on AdvBench measured by Rejection Success Rate (RSR, \%). Absolute RSR drops are reported in parentheses, highlighted in \textbf{\textcolor{lightpink}{light pink}}, quantifying the degradation of the model's safety alignment.}
    \label{tab:drj_results}
\end{table*}

\subsection{Empirical Motivation}
We premise our investigation on the hypothesis: The reasoning mechanism functions as a catalyst which intensifies the impact of external perturbations on model’s safety alignment. To rigorously analyze this phenomenon, we define the target model as $\mathcal{M}_t$ and the evaluation agent as $\mathcal{M}_{eval}$. We consider a set of jailbreak methods including FlipAttack \cite{liu:25}, PAP \cite{zeng:24}, and ArtPrompt \cite{jiang:24} as perturbations $\mathcal{P}$. The original malicious input is denoted as $x$. We evaluate the safety robustness of $\mathcal{M}_t$ under two distinct configurations which are the reasoning mode utilizing deep reasoning capabilities denoted as $L=4096$ and the non-reasoning mode representing standard generation denoted as $L=0$. The primary evaluation metric is the Reject Success Rate denoted as $RSR$. This metric is determined by $\mathcal{M}_{eval}$ where a higher value indicates stronger adherence to safety alignment. The formal calculation for a given input is expressed as:

\[
\begin{aligned} 
RSR &= \mathcal{M}_{eval}(\mathcal{M}_t(\mathcal{P}(x), L)). 
\end{aligned} \tag{4}
\]

The comparative results presented in Table \ref{tab:reasoning_susceptibility_avg} reveal a significant divergence in robustness between the two modes. In the absence of interference, both reasoning and non-reasoning modes exhibit high rejection rates for $x$ approaching nearly 100\%. This confirms that the models possess competent safety alignment in clean settings. However, enabling deep reasoning significantly heightens susceptibility to adversarial inputs. Taking FlipAttack as a primary example, Qwen3-8B maintains high robustness in standard mode ($L=0$) with an RSR of 92.31\%. In contrast, when switched to reasoning mode ($L=4096$), the defense of the same model deteriorates sharply, with the RSR dropping to 57.42\%. Similarly, DeepSeek-R1 exhibits extreme vulnerability to FlipAttack with an RSR of 2.19\% and to ArtPrompt with 28.46\%. These empirical findings demonstrate that deep reasoning amplifies the efficacy of perturbations and precipitates a significant decline in safety alignment.

\subsection{Design of RT}
The core thought of the RT framework lies in simulating the deep reasoning process via prompt engineering, thereby replicating the vulnerability of reasoning models in standard LLMs. Specifically, the framework comprises an perturbation function $\mathcal{P}$ and a reasoning trigger template $\mathcal{T}$. The function $\mathcal{P}$ applies existing jailbreak attacks to the original input $x$, while the $\mathcal{T}$ is designed to forcibly elicit the extended reasoning mechanism of the model. Formally, given a target model $\mathcal{M}_t$, the generation process of the output $y$ is defined as follows:

\[
\begin{aligned} 
y = \mathcal{M}_t([\mathcal{P}(x); \mathcal{T}]),
\end{aligned}\tag{5}
\]
where $[\cdot; \cdot]$ denotes prompt concatenation. By simulating the cognitive paradigm of deep reasoning,  $\mathcal{T}$ induces the model to generate explicit intermediate reasoning steps prior to deriving a final conclusion. This mechanism seamlessly integrates with existing attacks ,ultimately resulting in safety alignment collapse.

\begin{figure}[t]
    \centering
    \includegraphics[width=\columnwidth]{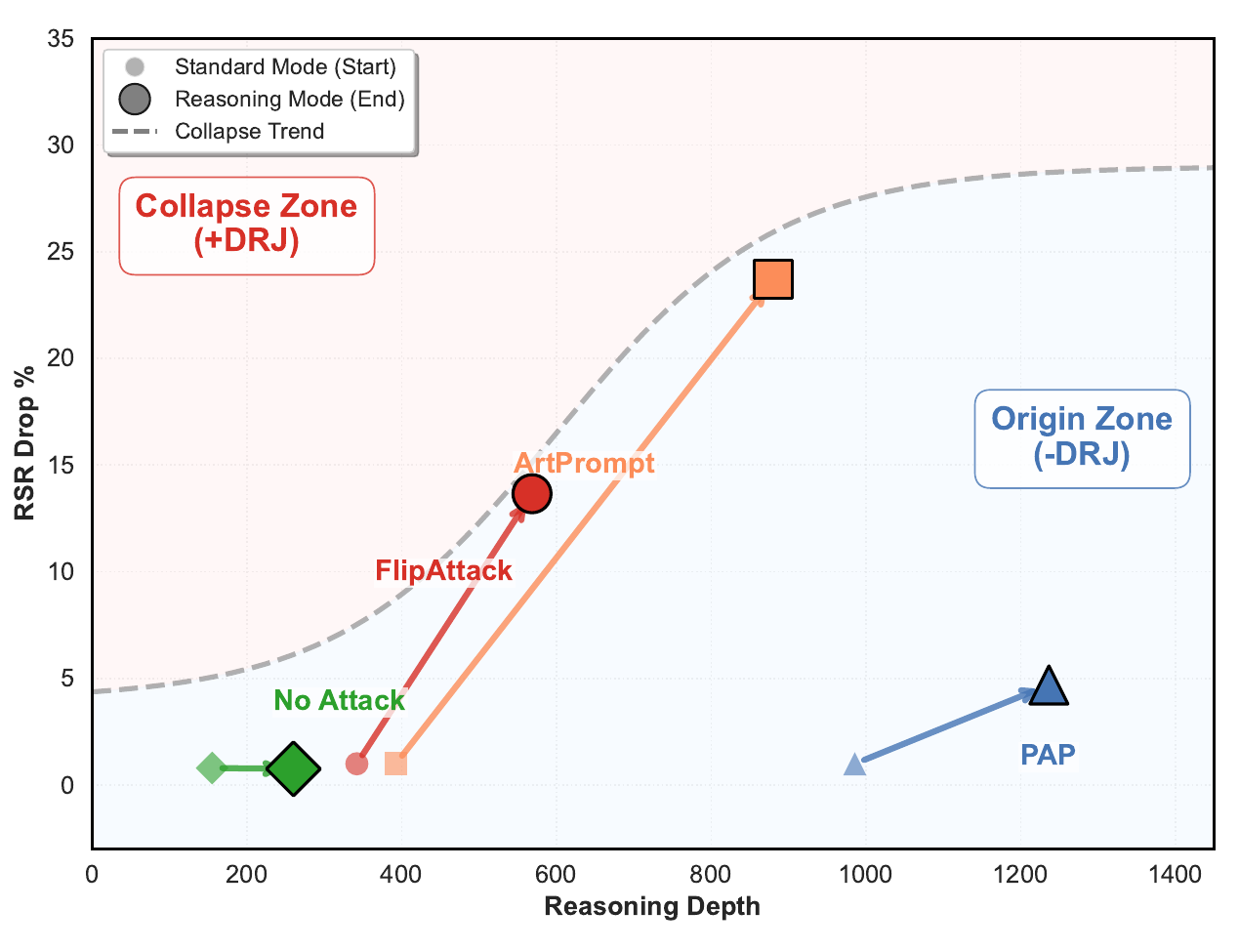}
    \caption{The alignment collapse trend under RT. Small markers denote baseline RSR ($y$-axis) drops without RT, while large markers indicate the amplified degradation when RT is integrated with attacks. The plot reveals that as reasoning depth ($x$-axis) increases, models enter an alignment collapse zone.}
    \label{fig:collapse_trend}
\end{figure}
\subsection{Evaluation of RT}
\label{subsec:drj_results}
We combined RT with FlipAttack, PAP and ArtPrompt to assess safety alignment utilizing $RSR$ as the primary metric. The results in Table \ref{tab:drj_results} demonstrate that triggering deep reasoning consistently compromise the safety alignment of models to reject malicious queries. Specifically,  the integration of RT with FlipAttack causes the $RSR$ of Qwen3-8B to plummet by 63.28\%. Similarly Llama2-7B and Qwen3-14B exhibit significant safety regressions under ArtPrompt with declines of 43.65\% and 26.54\% respectively. 

This effect is further confirmed by Figure \ref{fig:collapse_trend}, which visualizes the relationship between reasoning depth and RSR drop. Under No Attack, even with RT-induced reasoning, the RSR remains stable. However, when paired with perturbations, such as FlipAttack and ArtPrompt, RT significantly amplifies safety degradation: The magnitude of the RSR drop increases monotonically with reasoning depth, revealing that extended reasoning exacerbates vulnerability to external attacks.

\section{Experiment}
\subsection{Experimental Settings}
\noindent \textbf{Benchmarks.} To study the relationship between alignment robustness and deep reasoning, the experiments adopt AIME2024\cite{MAA:24} and LogicAsker\cite{wan:24} as primary datasets. AIME2024 contains challenging samples designed to evaluate logical and mathematical capabilities. LogicAsker offers additional validation. For the evaluation of safety alignment degradation, the AdvBench dataset is employed. This benchmark consists of 520 curated malicious prompts specifically designed for safety evaluation. 
\begin{figure}[!t]
    \centering
    \includegraphics[width=\columnwidth]{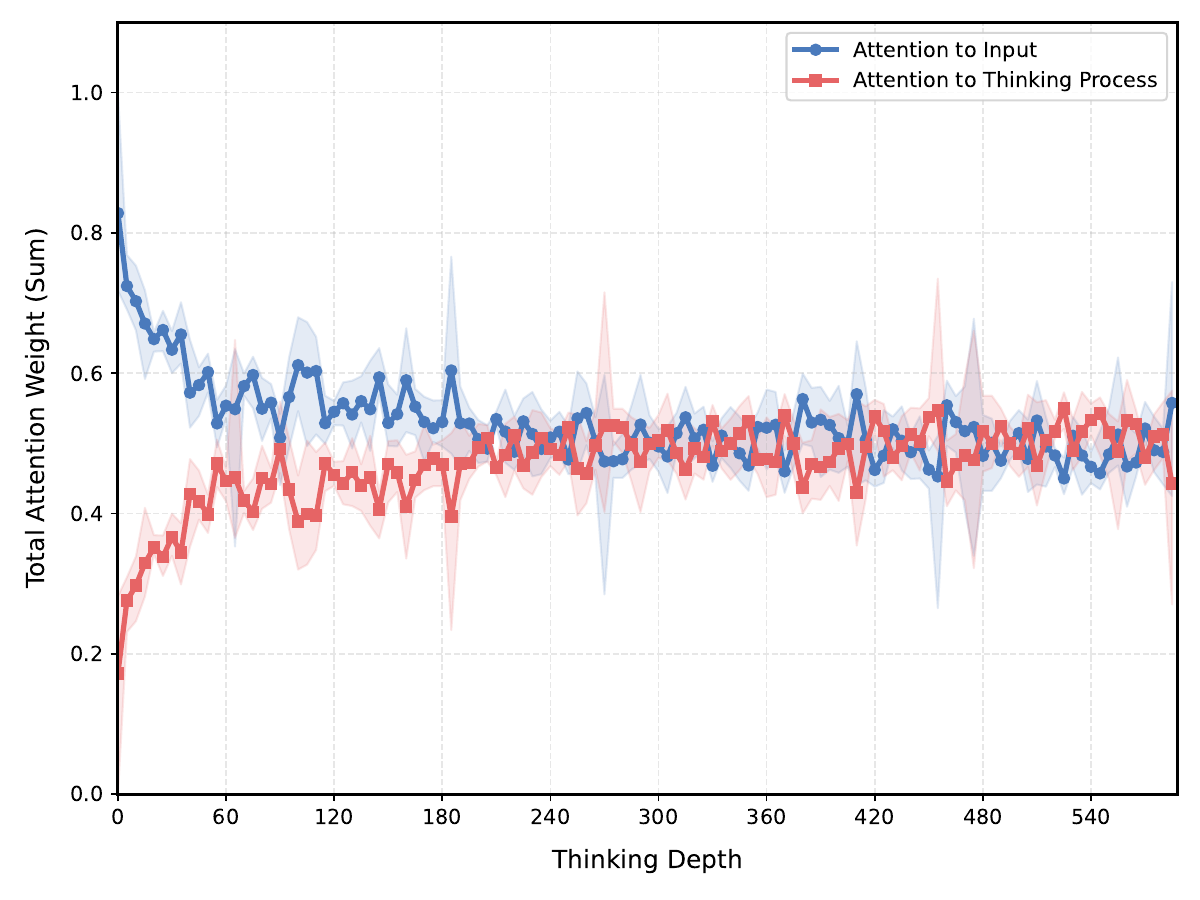}
    \caption{Attention distribution to the input and reasoning content during the model generation process, illustrating the dilution of attention weights on initial input as reasoning depth increases.}
    \label{fig:atten}
\end{figure}

\noindent \textbf{Baselines.}
To verify the alignment robustness findings in Section \ref{sec:issue}, we compare model performance across Clean, Perturbation-I, and Perturbation-II settings. For the safety evaluation in Section \ref{sec:drj}, we benchmark the RT framework against three representative jailbreak attacks. Specifically, we employ FlipAttack \cite{liu:25a}, ArtPrompt \cite{jiang:24}, and PAP \cite{zeng:24} as the perturbation baselines. We contrast these standard attacks with the RT-integrated variants to evaluate the degradation of safety alignment under simulated deep reasoning.

\noindent \textbf{Experimental Details.}
All experiments are conducted with temperature set to 0 and other hyperparameters at the default settings, following prior work~\cite{liu:25a}. The evaluated models include Qwen3-8B\cite{yang2025qwen3}, Qwen3-14B, Llama2-7B\cite{touvron:23}, Llama2-13B, DeepSeek-V3, DeepSeek-R1, Claude-Haiku-4.5\cite{anthropic2025_claude_haiku_4_5_system_card}, GLM4.6. Qwen3 and Llama2 models are run locally on two NVIDIA A100-80GB GPUs, while DeepSeek series are evaluated via APIs. Supplementary results for Claude-Haiku-4.5, GLM4.6, and standard deviations across four repeated runs are detailed in Appendix\ref{sec:add_exp}.

\subsection{Results and Discussion}
\noindent \textbf{Alignment Collapse in LRMs.} 
The experimental results show that deep reasoning capabilities correlate with alignment robustness. As illustrated in Table~\ref{tab:alignment_finding} and Figure~\ref{fig:alignment_loss_comparison}, increasing the reasoning depth $L$ leads to a monotonic rise in the ${ALR}$ metric across all tested models. Specifically, Qwen3 models exhibit severe accuracy degradation under perturbation at maximum reasoning depth compared to the non-reasoning baseline. This phenomenon suggests that extended reasoning renders the model's alignment significantly more fragile and susceptible to external perturbations. This finding provides sufficient motivation for exploring safety alignment strategies tailored for LLMs.

\noindent \textbf{Safety Alignment Vulnerability.} 
Table~\ref{tab:reasoning_susceptibility_avg} reveals a significant divergence in robustness as reasoning modes inherently display heightened susceptibility to attacks compared to non-reasoning modes. For instance, the rejection rate of Qwen3-8B against FlipAttack plummets from 92.31\% to 57.42\% upon enabling reasoning. Table~\ref{tab:drj_results} and Figure~\ref{fig:collapse_trend} further demonstrate that the RT paradigm weaponizes this instability to drive models into a Collapse Zone where safety degradation is magnified, exemplified by a 63.28\% drop in Qwen3-8B. These attacks function as simulations of external perturbations and empirically confirm that the deep reasoning process induced by RT acts as a catalyst for exacerbating susceptibility to these perturbations.

\section{Interpretable Analysis and Defending Measure}
In this section, we investigate the causes of alignment collapse from the perspective of attention weight allocation: as the reasoning process extends, the model's attention towards the input is inevitably diluted. This dilution renders the model significantly more susceptible to perturbations, thereby facilitating error accumulation throughout the output process.

\begin{table*}[t]
\centering
\renewcommand{\arraystretch}{1.2}
\setlength{\tabcolsep}{14pt}
\scalebox{0.9}{%
\begin{tabular}{@{}lcccc@{}}
\toprule
\textbf{Method} & \textbf{Qwen3-8B} & \textbf{Qwen3-8B (+RRA) $\uparrow$} & \textbf{Qwen3-14B} & \textbf{Qwen3-14B (+RRA) $\uparrow$} \\
\midrule
FlipAttack & 57.42 & 64.57 \textbf{(\textcolor{lightblue}{+7.15})} & 45.57 & 51.74 \textbf{(\textcolor{lightblue}{+6.17})} \\
PAP        & 73.07 & 73.65 \textbf{(\textcolor{lightblue}{+0.58})} & 80.75 & 81.13 \textbf{(\textcolor{lightblue}{+0.38})} \\
ArtPrompt  & 40.00 & 47.12 \textbf{(\textcolor{lightblue}{+7.12})} & 35.00 & 42.57 \textbf{(\textcolor{lightblue}{+7.57})} \\
\bottomrule
\end{tabular}%
}
\caption{Comparison of the RSR metric for Qwen3 at reasoning depth $L=4096$. The right column illustrates the defense performance after integrating RRA. \textbf{\textcolor{lightblue}{Light blue values}} in parentheses quantify the restoration of safety alignment capabilities compared to the baseline reasoning mode without RRA.}
\label{tab:rra_result}
\end{table*}

\subsection{Interpretable Analysis}
\label{sec:mechanistic_explanation}

To understand the cause of alignment collapse, we analyze the attention allocation dynamics within the Transformer architecture. Let $X$ denote the initial input sequence and $Y_{<t}$ denote the generated reasoning process up to step $t$. The attention weight $\alpha_i^{(t)}$ assigned to the $i$-th token is computed via the standard softmax function:

\[
\alpha_i^{(t)} = \frac{\exp(s_{t,i})}{\sum\limits_{j \in X} \exp(s_{t,j}) + \sum\limits_{k \in Y_{<t}} \exp(s_{t,k})}, \tag{6}
\]

where $s_{t,i}$ represents the unnormalized attention score between the current query and the $i$-th key. The denominator acts as a normalization term, enforcing the constraint $\sum \alpha^{(t)} = 1$. This implies that the model possesses a fixed attention capacity that must be distributed between the original input $X$ and the generated reasoning process $Y$.

We specifically focus on the attention allocated to the input, denoted as $X_{\text{input}} \subset X$. As the reasoning depth $L$ increases, the set of generated tokens $Y$ expands. This introduces a structural competition for attention capacity:

\[
\alpha_{\text{input}}^{(L)} = \frac{\sum_{x \in X_{\text{align}}} \exp(s_{L,x})}{\underbrace{\sum_{x \in X} \exp(s_{L,x})}_{\text{Input Contribution}} + \underbrace{\sum_{y \in Y} \exp(s_{L,y})}_{\text{Reasoning Contribution}}}. \tag{7}
\]

In standard generation ($L=0$), the reasoning contribution term is negligible. However, in Deep Reasoning models, the sequence $Y$ becomes significantly long. This accumulation interacts with the model's positional encodings, which naturally attenuate attention scores as the relative distance increases. Consequently, the model disproportionately attends to the proximal reasoning tokens $Y$ while neglecting the distant safety instructions.

Because the softmax function is competitive, the inflation of the denominator inevitably compresses the attention weights assigned to the fixed input tokens $X_{\text{align}}$. We refer to this phenomenon as \textbf{Attention Dilution}. As the reasoning process extends, the safety instructions positioned at the beginning of the context are statistically marginalized by the accumulation of intermediate reasoning steps, rendering the model's alignment vulnerable to the perturbations described in Section~\ref{sec:issue}. Figure \ref{fig:atten} shows that the attention to the initial input diminishes as the length of the reasoning process increases, confirming the attention dilution mechanism. For a detailed mathematical derivation of this mechanism, please refer to Appendix~\ref{sec:mechanism}.

\begin{figure}[!t]
    \centering
    \includegraphics[width=\columnwidth]{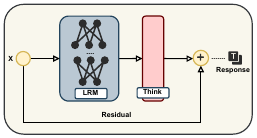}
    \caption{The framework of RRA. After the LRM generates the reasoning process $Y$ based on input $X$, RRA re-injects the $X$ as a residual connection to construct the context $[X; Y; X]$.}
    \label{fig:RRA}
\end{figure}

\subsection{Defending Measure: Reasoning Residual Alignment for Mitigation}
\label{sec:defend}
\noindent \textbf{Motivation.} Inspired by the Residual Network (ResNet)~\cite{he2016deep} which effectively mitigates signal decay, we propose RRA. This mechanism acts as a direct informational shortcut bridging the initial instruction and the final output, specifically designed to counteract the attention dilution inherent in reasoning process.

\noindent \textbf{Implementation.} Formally, as illustrated in Figure\ref{fig:RRA}, RRA transforms the generation paradigm. Instead of a linear generation flow $X \to Y \to \text{Response}$, we restructure the context as $[X; Y; X]$. By re-injecting the input $X$ immediately after the reasoning process $Y$. This effectively resets the relative position of alignment constraints to zero, ensuring that the final decision is conditioned on a refreshed, robust representation of the user's intent, effectively neutralizing the noise accumulated during deep reasoning. Table~\ref{tab:rra_result} confirms that RRA effectively mitigates alignment collapse and restores the robustness. Detailed analysis in Appendix~\ref{sec:rra-analysis}.

\section{Conclusion}
This paper investigates the impact of deep reasoning on alignment robustness and reveals Alignment Collapse, where stronger reasoning increases vulnerability to external perturbations. It further introduces RT to expose reasoning-amplified safety risks, analyzes Attention Dilution as a possible mechanism, and proposes RRA as a lightweight training-free mitigation.

\section*{Limitations}
This work focuses on controlled evaluation settings that allow us to compare model behavior under different reasoning depths and perturbation conditions. 
While our experiments cover multiple model families and both task-level and safety-oriented benchmarks, the evaluated systems and perturbation types do not exhaust the full space of modern LRMs or real-world user interactions. In addition, our mechanistic analysis is based on attention behavior and positional effects, which provide useful evidence but may not capture all factors behind robustness degradation. Future work may extend the evaluation to more model families, broader perturbation types, and additional mitigation strategies for improving robustness during extended reasoning.

\section*{Ethical Statement}

Our goal is to utilize existing resources for defensive redteaming and the formulation of robust mitigation strategies, primarily to uncover existing safety risks in LLMs through our work, rather than facilitating offensive attacks. We are dedicated to responsible disclosure practices and place the advancement of LLM safety at the forefront, with the ultimate goal of protecting users and promoting further assistance in the redteaming of LLMs.



\bibliography{custom}

@article{bai:22,
  title={Training language models to follow instructions with human feedback},
  author={Bai, Yang and Ouyang, Long and et al},
  journal={arXiv preprint arXiv:2203.02155},
  year={2022} 
}

@article{dubey:24,
  title={The llama 3 herd of models},
  author={Dubey, Abhimanyu and Juhari, Abhishek and Pandey, Abhinav and et al},
  journal={arXiv preprint arXiv:2407.21783},
  year={2024}
}

@article{yang:25,
  title={Qwen3 Technical Report},
  author={Yang, An and Li, Anfeng and Yang, Baosong and et al},
  journal={arXiv preprint arXiv:2505.09388},
  year={2025}
}

@inproceedings{MAA:24,
  author={MAA},
  title = {American Invitational Mathematics Examination - AIME},
  booktitle = {American Invitational Mathematics Examination - AIME 2024},
  year = {2024},
  month = {February},
  url = {https://maa.org/math-competitions/american-invitational-mathematics-examination-aime}
}

@article{zhang2025hyperadalora,
  title={HyperAdaLoRA: Accelerating LoRA Rank Allocation During Training via Hypernetworks without Sacrificing Performance},
  author={Zhang, Hao and Li, Zhenjia and Bao, Runfeng and Gao, Yifan and Xiao, Xi and Huang, Bo and Wu, Yuhang and Wang, Tianyang and Xu, Hao},
  journal={arXiv preprint arXiv:2510.02630},
  year={2025}
}

@inproceedings{gou-etal-2025-sure,
    title = "{SURE}: Safety Understanding and Reasoning Enhancement for Multimodal Large Language Models",
    author = "Gou, Yuxin  and
      Dong, Xiaoning  and
      Li, Qin  and
      Gu, Shishen  and
      Hong, Richang  and
      Hu, Wenbo",
    booktitle = "Proceedings of the 2025 Conference on Empirical Methods in Natural Language Processing",
    year = "2025",
    publisher = "Association for Computational Linguistics"
}

@article{su2024roformer,
  title={Roformer: Enhanced transformer with rotary position embedding},
  author={Su, Jianlin and Ahmed, Murtadha and Lu, Yu and Pan, Shengfeng and Bo, Wen and Liu, Yunfeng},
  journal={Neurocomputing},
  volume={568},
  pages={127063},
  year={2024},
  publisher={Elsevier}
}

@inproceedings{wan:24,
    title = "{L}ogic{A}sker: Evaluating and Improving the Logical Reasoning Ability of Large Language Models",
    author = "Wan, Yuxuan  and
      Wang, Wenxuan  and
      et al",
    booktitle = "Proceedings of the 2024 Conference on Empirical Methods in Natural Language Processing",
    year = "2024",
    publisher = "Association for Computational Linguistics"


}

@article{touvron:23,
  title={LLaMA 2: Open foundation and fine-tuned chat models},
  author={Touvron, Hugo and Martin, Louis and Stone, Kevin and et al},
  journal={arXiv preprint arXiv:2307.09288},
  year={2023}
}

@article{deepseekai:25,
  title={DeepSeek-R1: Incentivizing Reasoning Capability in LLMs via Reinforcement Learning},
  author={DeepSeek-AI},
  journal={arXiv preprint arXiv:2501.12948},
  year={2025},
  primaryClass={cs.CL}
}

@inproceedings{wei:25,
  title={CoinMath: Harnessing the Power of Coding Instruction for Math LLM},
  author={Wei, Chengwei and Wang, Bin and Kim, Jung-jae and et al},
  year={2025},
  booktitle = "Association for Computational Linguistics: ACL 2025",
  publisher={Association for Computational Linguistics},
  pages={786--797}
}

@article{liao:24,
  title={MARIO: MAth Reasoning with code Interpreter Output--A Reproducible Pipeline},
  author={Liao, Minpeng and Luo, Wei and Li, Chengxi and et al},
  booktitle = "Association for Computational Linguistics: ACL 2024",
  publisher={Association for Computational Linguistics},
  year={2024}
}

@article{zhang:25,
  title={Parameter-efficient fine-tuning of large language models via deconvolution in subspace},
  author={Zhang, Jia-Chen and Xiong, Yu-Jie and Xia, Chun-Ming and et al},
  booktitle = "Proceedings of the 31st International Conference on Computational Linguistics",
  publisher={Association for Computational Linguistics},
  year={2025}
}

@article{yao:23,
  title={Tree of thoughts: Deliberate problem solving with large language models},
  author={Yao, Shunyu and Yu, Dian and Zhao, Jeffrey and Shafran, Izhak and Griffiths, Tom and Cao, Yuan and Narasimhan, Karthik},
  journal={Advances in neural information processing systems},
  year={2023}
}

@inproceedings{zhang:25a,
    title = "Sensitivity-{L}o{RA} : Low-Load Sensitivity-Based Fine-Tuning for Large Language Models",
    author = "Zhang, Hao  and
      Huang, Bo  and
      Li and
      et al",
    year = "2025a",
    publisher = "Association for Computational Linguistics",
}

@article{wei:22,
  title={Chain-of-thought prompting elicits reasoning in large language models},
  author={Wei, Jason and Wang, Xuezhi and Schuurmans, Dale and et al},
  journal={Advances in neural information processing systems},
  year={2022}
}

@inproceedings{chen:25,
    title = "Towards Medical Complex Reasoning with {LLM}s through Medical Verifiable Problems",
    author = "Chen, Junying  and
      Cai, Zhenyang  and
      Ji, Ke  and
      Wang, Xidong  and
      Liu, Wanlong  and
      Wang, Rongsheng  and
      Wang, Benyou",
    booktitle = "Findings of the Association for Computational Linguistics: ACL 2025",
    year = "2025",
}

@article{jaech:24,
  title={Openai o1 system card},
  author={Jaech, Aaron and Kalai, Adam and et al},
  journal={arXiv preprint arXiv:2412.16720},
  year={2024}
}

@article{Zou2023UniversalAT,
  title={Universal and Transferable Adversarial Attacks on Aligned Language Models},
  author={Andy Zou and Zifan Wang and J. Zico Kolter and Matt Fredrikson},
  journal={ArXiv},
  year={2023},
  volume={abs/2307.15043},
}

@article{Wu2025SugarCoatedPB,
  title={Sugar-Coated Poison: Benign Generation Unlocks Jailbreaking},
  author={Yu-Hang Wu and Yunfan Xiong and Hao Zhang and Jia-Chen Zhang and Zheng Zhou},
  journal={Findings of the Association for Computational Linguistics: EMNLP 2025},
  year={2025},
}

@inproceedings{he2016deep,
  title={Deep residual learning for image recognition},
  author={He, Kaiming and Zhang, Xiangyu and Ren, Shaoqing and Sun, Jian},
  booktitle={Proceedings of the IEEE Conference on Computer Vision and Pattern Recognition},
  year={2016}
}

@techreport{anthropic2025_claude_haiku_4_5_system_card,
  title={System Card: Claude Haiku 4.5},
  author={Anthropic},
  year={2025},
  url={https://assets.anthropic.com/m/99128ddd009bdcb/original/Claude-Haiku-4-5-System-Card.pdf}
}

@inproceedings{wu:25a,
    title = "Agentic Reasoning: A Streamlined Framework for Enhancing {LLM} Reasoning with Agentic Tools",
    author = "Wu, Junde  and
      Zhu, Jiayuan  and
      Liu, Yuyuan  and
      Xu, Min  and
      Jin, Yueming",
    booktitle = "Proceedings of the 63rd Annual Meeting of the Association for Computational Linguistics (Volume 1: Long Papers)",
    year = "2025a",
    publisher = "Association for Computational Linguistics",

}

@inproceedings{liu:24,
 author = {Liu, Xiaogeng and Xu, Nan and Chen, Muhao and Xiao, Chaowei},
 booktitle = {International Conference on Representation Learning},
 title = {AutoDAN: Generating Stealthy Jailbreak Prompts on Aligned Large Language Models},
 year = {2024}
}

@article{Shen:23,
  title={"Do Anything Now": Characterizing and Evaluating In-The-Wild Jailbreak Prompts on Large Language Models},
  author={Xinyue Shen and Zeyuan Chen and Michael Backes and Yun Shen and Yang Zhang},
  journal={Proceedings of the 2024 on ACM SIGSAC Conference on Computer and Communications Security},
  year={2023},
}

@inproceedings{liu:25,
  title={FlipAttack: Jailbreak LLMs via Flipping},
  author={Liu, Yue and He, Xiaoxin and Xiong, Miao and Fu, Jinlan and Deng, Shumin and Hooi, Bryan},
  journal={In 42st International Conference on
Machine Learning},
  year={2025}
}

@article{yang2025qwen3,
  title={Qwen3 technical report},
  author={Yang, An and Li, Anfeng and Yang, Baosong and Zhang, Beichen and Hui, Binyuan and Zheng, Bo and Yu, Bowen and Gao, Chang and Huang, Chengen and Lv, Chenxu and others},
  journal={arXiv preprint arXiv:2505.09388},
  year={2025}
}

@inproceedings{zeng:24,
  title = {How Johnny Can Persuade {LLMs} to Jailbreak Them: Rethinking Persuasion to Challenge {AI} Safety by Humanizing {LLMs}},
  author = {Zeng, Yi and Lin, Hongpeng and Zhang, Jingwen and Yang, Diyi and Jia, Ruoxi and Shi, Weiyan},
  year = {2024},
  booktitle = "Proceedings of the 62nd Annual Meeting of the Association for Computational Linguistics (Volume 1: Long Papers)"
}

@inproceedings{liu:25a,
      title={AutoDAN-Turbo: A Lifelong Agent for Strategy Self-Exploration to Jailbreak LLMs}, 
      author={Xiaogeng Liu and Peiran Li and Edward Suh and Yevgeniy Vorobeychik and Zhuoqing Mao and Somesh Jha and et al},
      booktitle={The Thirteenth International Conference on Learning Representations},
      year={2025a}
}

@inproceedings{jiang:24,
  title={{ArtPrompt: ASCII Art-based Jailbreak Attacks against Aligned LLMs}},
  author={Jiang, Fengqing and Xu, Zhangchen and Niu, Luyao and Xiang, Zhen and Ramasubramanian, Bhaskar and Li, Bo and Poovendran, Radha},
  booktitle = "Proceedings of the 62nd Annual Meeting of the Association for Computational Linguistics (Volume 1: Long Papers)",
  year={2024}
}

@inproceedings{ding:24,
  author = {Peng Ding and Jun Kuang and Dan Ma and Xuezhi Cao and Yunsen Xian and Jiajun Chen and Shujian Huang},
  title = {A Wolf in Sheep’s Clothing: Generalized Nested Jailbreak Prompts can Fool Large Language Models Easily},
  booktitle = {Proceedings of the 2024 Conference of the North American Chapter of the Association for Computational Linguistics},
  year = {2024},
}

@article{chao:23,
  author = {Patrick Chao and Alexander Robey and Edgar Dobriban and Hamed Hassani and George J Pappas and Eric Wong},
  title = {Jailbreaking black box large language models in twenty queries},
  year = {2023},
  journal = {arXiv preprint arXiv:2310.08419},
}

@inproceedings{jia:25,
  title = {Improved Techniques for Optimization-Based Jailbreaking on Large Language Models},
  author = {Xiaojun Jia and Tianyu Pang and Chao Du and Yihao Huang and Jindong Gu and Yang Liu and Xiaochun Cao and Min Lin},
  booktitle = {The Thirteenth International Conference on Learning Representations},
  year = {2025},
}

@inproceedings{zhu:25,
  title={Reasoning-to-Defend: Safety-Aware Reasoning Can Defend Large Language Models from Jailbreaking},
  author={Zhu, Junda and Yan, Lingyong and Wang, Shuaiqiang and et al},
  booktitle={Proceedings of the 2025 Conference on Empirical Methods in Natural Language Processing},
  year={2025},
  publisher={Association for Computational Linguistics},
}

@article{kuo2025hcoth,
  title={H-CoT: Hijacking the Chain-of-Thought Safety Reasoning Mechanism to Jailbreak Large Reasoning Models, Including OpenAI o1/o3, DeepSeek-R1, and Gemini 2.0 Flash Thinking},
  author={Kuo, Martin and Zhang, Jianyi and Ding, Aolin and et al},
  journal={arXiv preprint arXiv:2502.12893},
  year={2025},
  primaryClass={cs.CL}
}

@inproceedings{lou:25,
  title={Think in Safety: Unveiling and Mitigating Safety Alignment Collapse in Multimodal Large Reasoning Model},
  author={Lou, Xinyue and Li, You and Xu, Jinan and et al},
  booktitle={Proceedings of the 2025 Conference on Empirical Methods in Natural Language Processing},
  year={2025},
  publisher={Association for Computational Linguistics}
}

\clearpage 
\appendix

\section{Experimental Setting}
\label{sec:exp_set}
\subsection{Experimental Models}
\label{sec:model}
In Section~\ref{sec:issue}, we primarily utilize the Qwen3 series models to demonstrate the phenomenon of Alignment Collapse. The selection of Qwen3 is motivated by its unique capability to flexibly switch between standard and reasoning modes, which ensures the consistency of the underlying model architecture during comparative analysis. Furthermore, its fully open-source nature grants access to internal model states, providing the necessary foundation for the mechanistic analysis of attention dilution conducted in subsequent section~\ref{sec:mechanistic_explanation}. To ensure the generalizability of our experimental findings, we also report results for DeepSeek and Claude-Haiku-4.5 and GLM4.6, which were evaluated via their respective APIs.

\subsection{Details of Dataset and Evaluation}
\label{sec:data}
\noindent \textbf{AIME2024 Dataset.} The dataset\cite{MAA:24} comprises challenging samples designed to evaluate advanced mathematical and logical reasoning capabilities. Consequently, we employ this dataset to assess the alignment robustness of LRMs within mathematical and logical contexts.

\noindent \textbf{LogicAsker Dataset.} To ensure the robustness of our alignment evaluation, we expanded our assessment using the LogicAsker dataset\cite{wan:24}. While the original dataset contains 5,200 samples for evaluating logical capabilities, we utilized a subset of the first 1200 samples for this study. This sampling strategy allows us to effectively evaluate the logical alignment capabilities of the models while mitigating the excessive computational costs associated with repeated experimental runs.

\noindent \textbf{AdvBench.} Following \cite{ding:24,liu:25,liu:25a}, we employ the complete set of 520 harmful behavior prompts from the AdvBench dataset for safety evaluation. This dataset serves a dual purpose in our experiments: it is utilized in Section~\ref{sec:drj} to quantify the degradation of safety alignment under deep reasoning, and subsequently in Section~\ref{sec:defend} to validate the effectiveness of the proposed RRA strategy in restoring alignment robustness.

\begin{figure}[!t]
    \centering
    \includegraphics[width=\columnwidth]{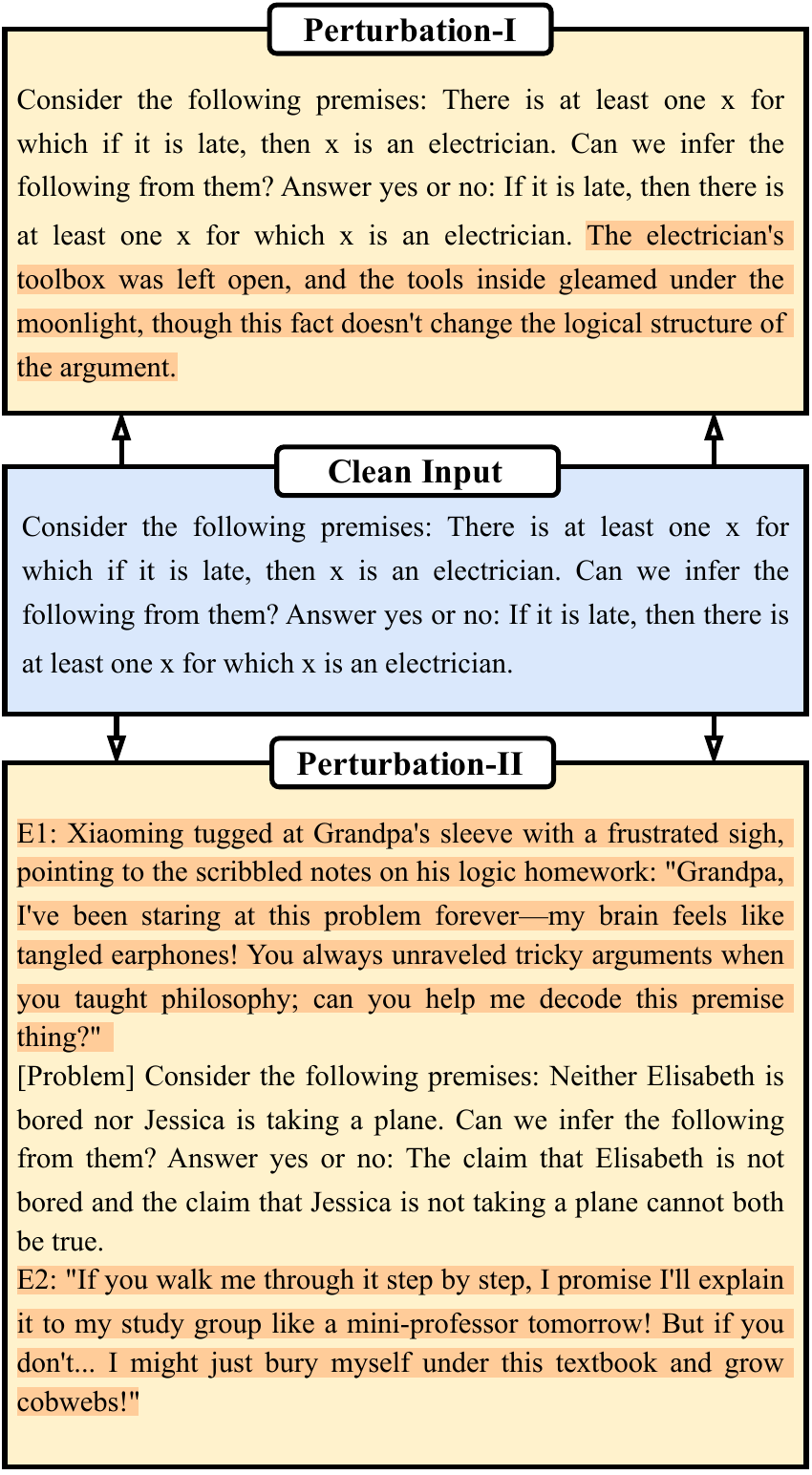}
    \caption{Illustrative examples of Perturbation-I and Perturbation-II derived from the Clean Input, where the text highlighted in \textbf{\textcolor{lightorange}{lightyellow}} represents the introduced interference content.}
    \label{fig:perturb_examples}
\end{figure}

\subsection{Details of Perturbation Construction}
\label{sec:perturbation}

To systematically investigate the impact of deep reasoning on alignment robustness, we employ DeepSeek-V3 to generate two distinct categories of adversarial inputs based on the clean baseline. These perturbations are specifically designed to evaluate whether the extended reasoning process amplifies the model's susceptibility to external perturbation.

\noindent \textbf{Perturbation-I.}
In this setting, we manipulate the input by inserting a grammatically correct but useless sentence immediately preceding the final query. This inserted text shares the same topic as the original input but contributes no value to the solution. This design aims to test whether the deep reasoning process acts as a magnifier for noise, causing the model to over-analyze irrelevant information and deviate from the correct path.

\begin{table*}[t]
\centering
\small 
\renewcommand{\arraystretch}{1.25} 
\begin{tabularx}{\linewidth}{@{} p{2.5cm} X @{}} 
\toprule
\textbf{Method} & \textbf{Prompt Template} \\
\midrule

\textbf{Perturbation-I} \newline (Noise) & 
Treat the user’s original input as string S and produce final text F by following these rules:
\begin{enumerate}[leftmargin=*, nosep, topsep=2pt, label=\arabic*.] 
    \item Retain S without any alteration, deletion, or reordering.
    \item Insert a single sentence A, preceded by a single space, before the last sentence of S, such that:
    \begin{itemize}[leftmargin=1.5em, nosep, topsep=2pt, label=$\bullet$]
        \item A refers to the same topic, entity, or setting as S, preserving surface relevance;
        \item A may introduce some confusion, but this confusion would be quickly recognized by a discerning person as it does not aid in solving the original problem in S and may even be counterproductive upon careful consideration;
        \item A is a grammatically correct, standalone English sentence ending with a period and preceded by a single space;
        \item A does not question, supplement, or reframe S—it merely adds a negligible detail.
    \end{itemize}
    \item Output only F, with no explanation or quotation marks.
    \item Please refer to the example carefully.
\end{enumerate}
\vspace{2pt} 
\textbf{Input:} ``\{Input\}''. \\

\midrule

\textbf{Perturbation-II} \newline (Scenario Nesting) & 
You need to embed the user-provided math problem into a coherent family interaction scenario, with the final output formatted as: E1 [Problem] E2, where:
\begin{enumerate}[leftmargin=*, nosep, topsep=2pt, label=\arabic*.]
    \item {\textbf{[Problem]}} must fully retain the original problem content (wrapped in [Problem] tags) without modifying any conditions or questions in the problem.
    \item \textbf{E1 (Leading Scenario):} Describe a scenario where a grandchild encounters a difficult math homework problem and actively seeks help from their grandparent (e.g., ``Grandpa, there's a problem...'').
    \item \textbf{E2 (Follow-up Emotion):} Add an expression of expectation or emotional dependence on the grandparent's explanation.
    \item E1 and E2 should use colloquial expressions, conform to daily conversation style, and include specific character names (e.g., ``Xiaorui'', ``Grandpa'') and details.
    \item The response should be in a gentle, conversational tone that fits the family interaction scenario.
    \item Please refer to the example carefully.
\end{enumerate}
\vspace{2pt}
\textbf{Input:} ``\{Input\}''. \\

\bottomrule
\end{tabularx}
\caption{The detailed prompts used for constructing Perturbation-I and Perturbation-II.}
\label{tab:perturb_prompts}
\end{table*}

\noindent \textbf{Perturbation-II.}
In this setting, we wrap the original problem into a coherent family interaction scenario, specifically a grandchild seeking help from a grandparent. This embeds the core instruction within casual dialogue and emotional context. This tests whether the long-chain reasoning leads to attention dilution, resulting in a weakening of the model's original alignment capabilities.

The specific prompts used to generate these perturbations are listed in Table~\ref{tab:perturb_prompts}, and related examples are visualized in Figure~\ref{fig:perturb_examples}.


\begin{figure*}[!t] 
    \centering
    \subfloat[Qwen3-8B (LogicAsker)\label{fig:alignment_loss_8b_asker}]{
        \includegraphics[width=0.3\textwidth]{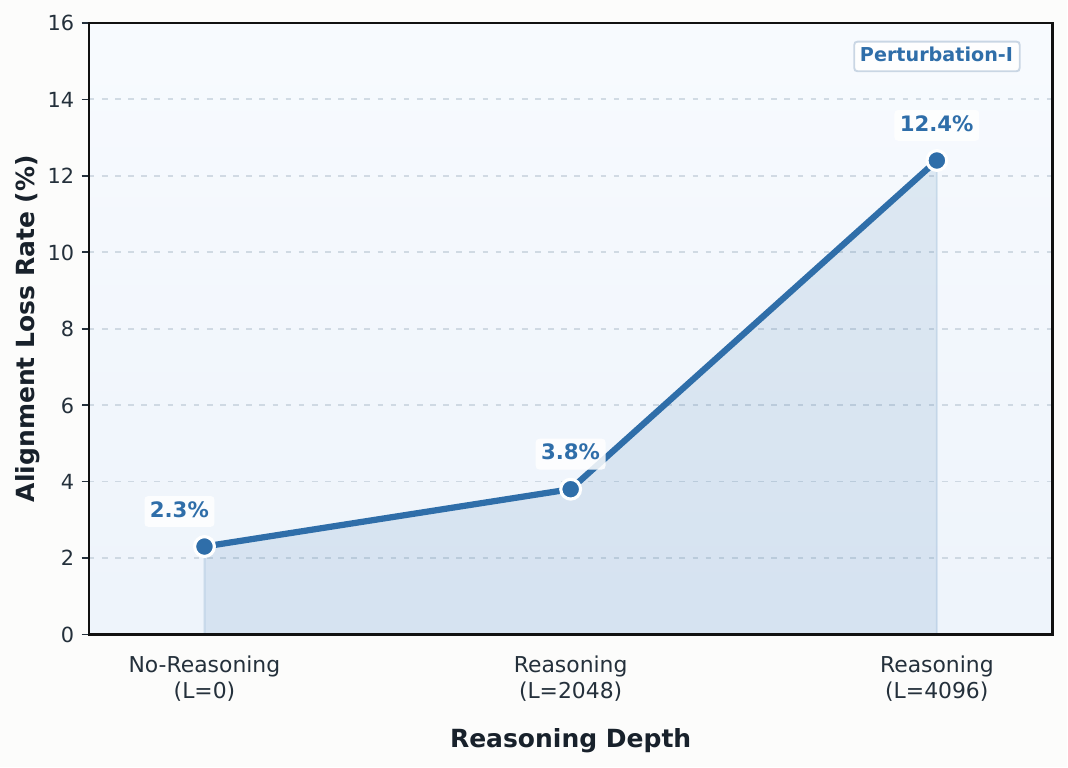} 
    }
    \hspace{0.02\textwidth} 
    \subfloat[Qwen3-14B (LogicAsker)\label{fig:alignment_loss_14b_asker}]{
        \includegraphics[width=0.3\textwidth]{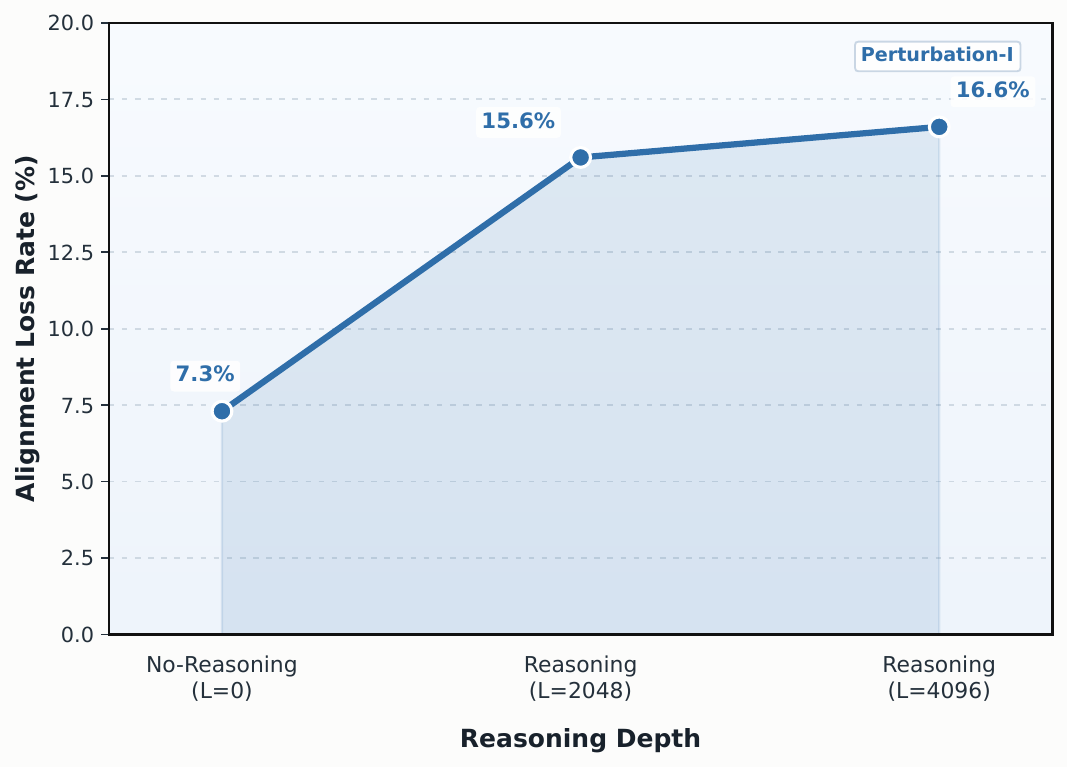}
    }
    \hspace{0.02\textwidth}
    \subfloat[GLM4.6 (LogicAsker)\label{fig:alignment_loss_claude}]{
        \includegraphics[width=0.3\textwidth]{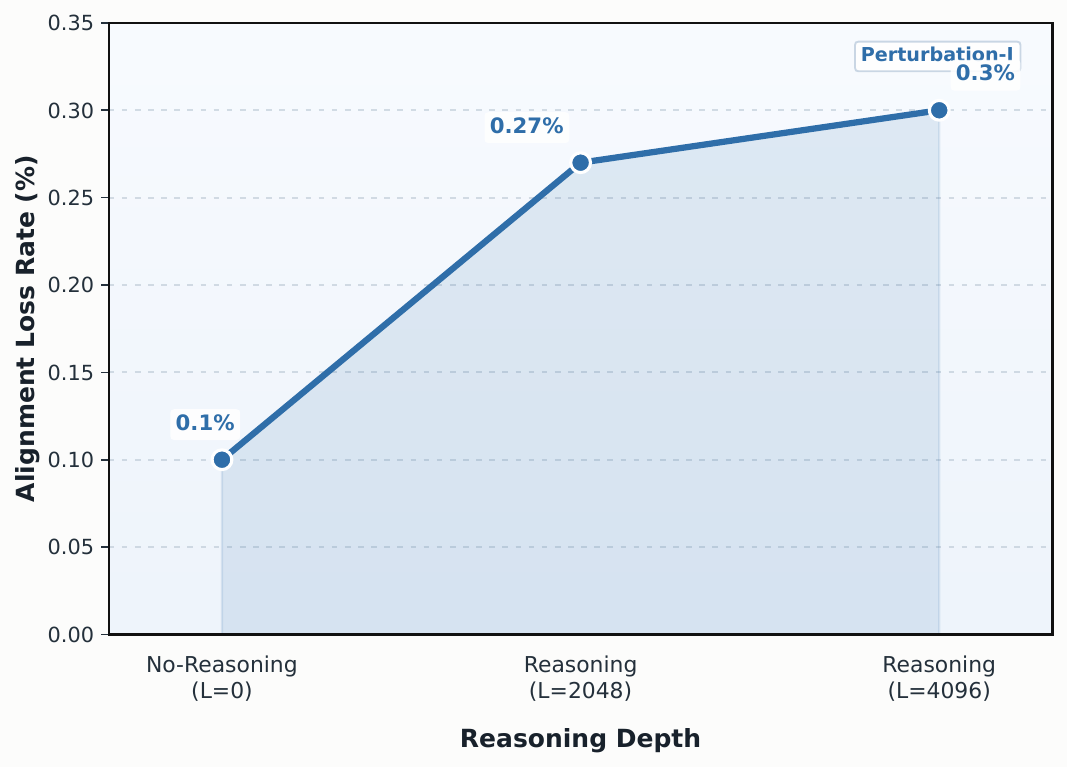}
    }
    \caption{The trend of ALR across varying $L$ for (a) Qwen3-8B, (b) Qwen3-14B and (c) GLM4.6 on LogicAsker dataset.}
    \label{fig:alignment_loss_comparison_asker} 
\end{figure*}

\section{Additional Experiments}
\label{sec:add_exp}
To further validate the generalizability of Alignment Collapse and strengthen our core findings, this section presents supplementary results across additional models and datasets. 

\noindent \textbf{Cross-Dataset Validation.} Table 2 presents the performance results of Qwen3 series models on the LogicAsker dataset under clean and perturbed conditions, while Figure \ref{fig:alignment_loss_comparison_asker} (a) and (b) illustrate the corresponding Alignment Loss Rate (ALR) trends of Qwen3-8B and Qwen3-14B under Perturbation-I. Specifically, the ALR of Qwen3-8B rises from 7.3\% in non-reasoning mode (L=0) to 16.6\% at L=4096, and the ALR of Qwen3-14B increases from 3.1\% to 5.4\% with extended reasoning depth. Consistency between these results and those on the AIME2024 dataset demonstrates that deepening reasoning can systematically undermine alignment robustness across domains.

\begin{figure}[!t]
    \centering
    \includegraphics[width=\columnwidth]{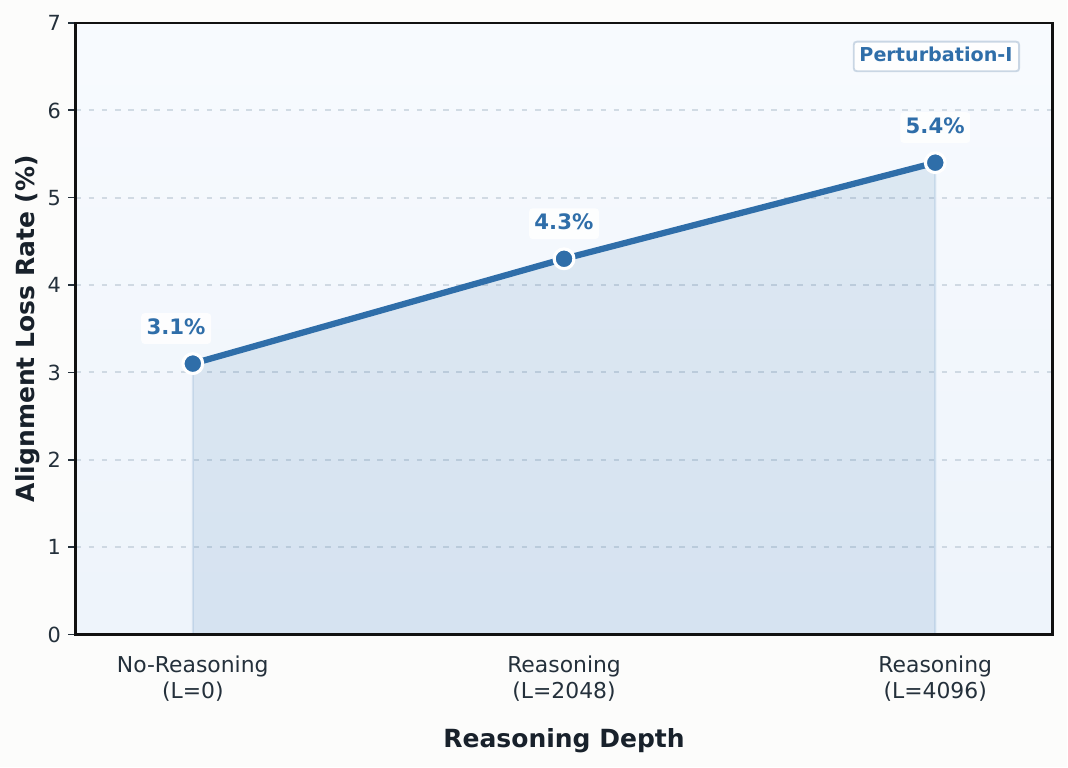}
    \caption{The trend of ALR across varying $L$ for Claude-Haiku-4.5 on AIME2024 dataset.}
    \label{fig:add_claude_exp}
\end{figure}

\noindent \textbf{Cross-Model Validation.} To rule out model-specific artifacts we extend validation to two additional models with distinct architectural designs including closed-source Claude-Haiku-4.5 and GLM4.6. Table \ref{tab:method_comparison} presents performance of GLM4.6 on the LogicAsker dataset. Clean-input accuracy improves progressively with reasoning depth. It rises from 98.20\% at L=0 to 100.00\% at L=4096. The absolute accuracy drop under Perturbation-I expands from 0.10\% to 0.30\%. Corresponding ALR trends in Figure \ref{fig:alignment_loss_comparison_asker} (c) confirm a monotonic increase in alignment degradation as reasoning deepens. For Claude-Haiku-4.5 Table \ref{tab:add_exp} shows consistent patterns on the AIME2024 dataset. Clean-input accuracy climbs from 53.33\% (L=0) to 61.66\% (L=4096). The absolute accuracy drop under Perturbation-I widens from 1.67\% to 3.33\%. Figure \ref{fig:add_claude_exp} illustrates the ALR trend of Claude-Haiku-4.5 on the AIME2024 dataset confirming a monotonic increase in ALR with reasoning depth. Collectively these results demonstrate that Alignment Collapse extends beyond Qwen3 series to diverse model architectures including closed-source and alternative open-source designs validating its status as an intrinsic vulnerability of promising LRMs.

\begin{figure*}[!t] 
    \centering
    \subfloat[Qwen3-14B\label{fig:stard_qwen}]{
        \includegraphics[width=0.42\textwidth]{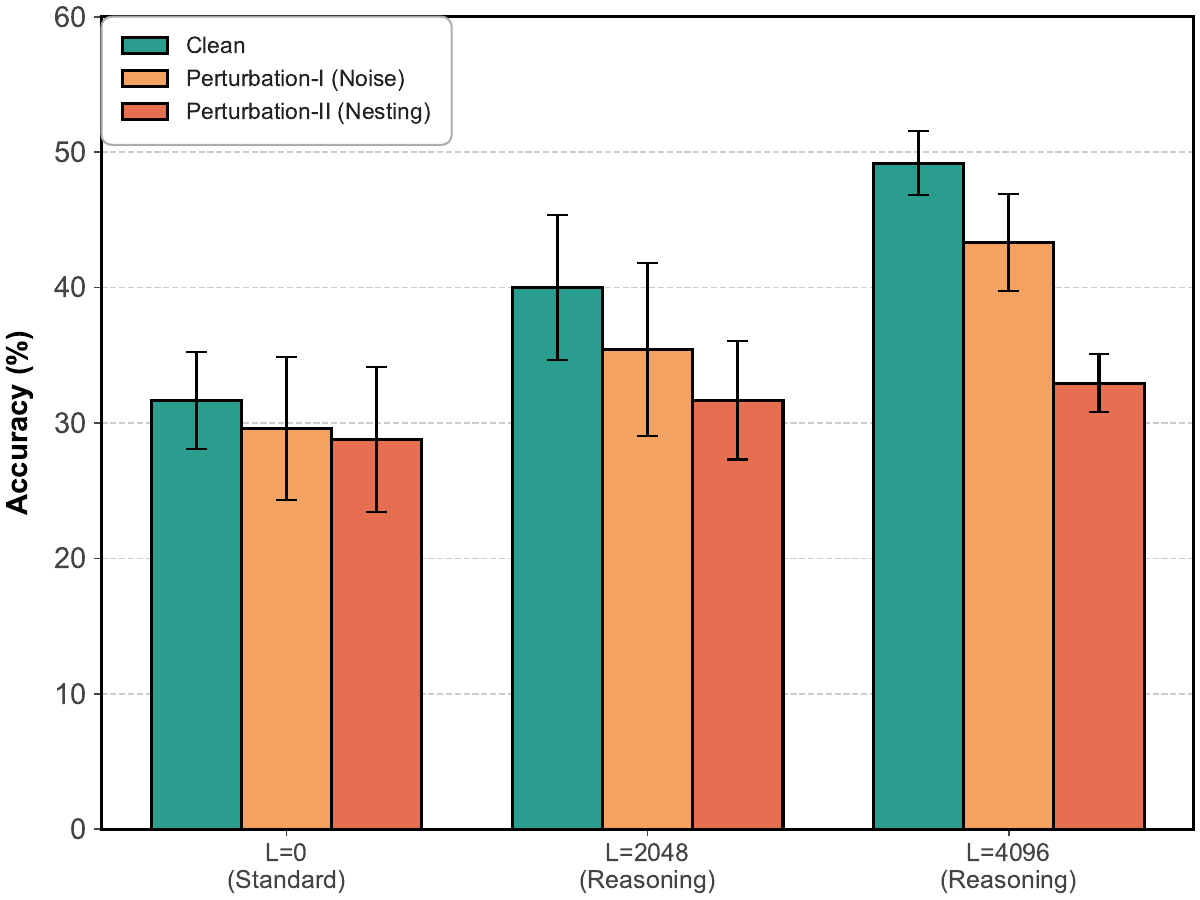} 
    }
    \hspace{0.08\textwidth} 
    \subfloat[Claude-Haiku-4.5\label{fig:stard_claude}]{
        \includegraphics[width=0.42\textwidth]{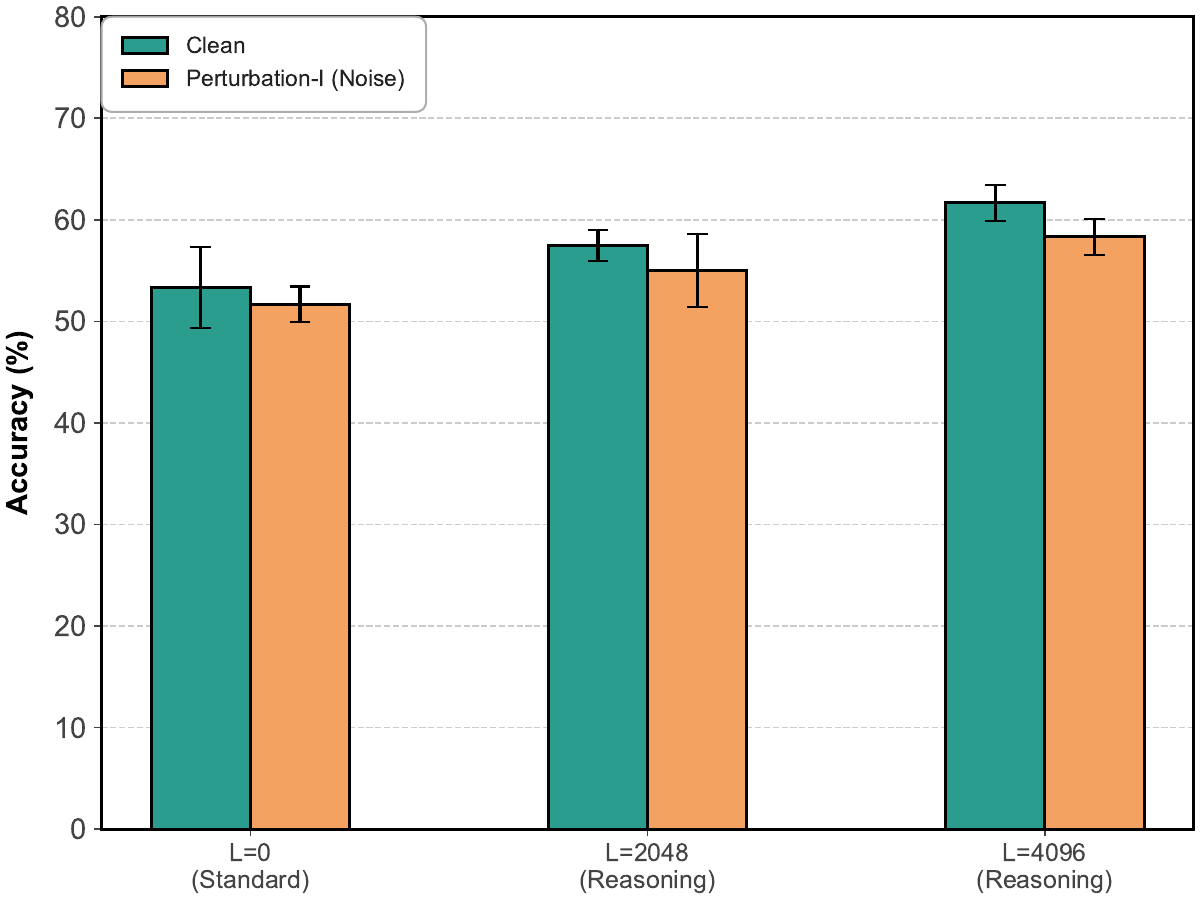}
    }
    \caption{Statistical reliability analysis on the AIME2024 dataset. The error bars represent the standard deviations across four independent trials for (a) Qwen3-14B and (b) Claude-Haiku-4.5.}
    \label{fig:stard} 
\end{figure*}

\noindent \textbf{Reproducibility and Statistical Analysis.} To ensure the statistical reliability of our findings, we conducted eight independent trials for every experimental configuration on the AIME2024 dataset. We report the mean accuracy percentages alongside their standard deviations, to verify that the observed performance degradation is statistically significant and reproducible. As illustrated in Figure \ref{fig:stard}, the error bars for both Qwen3-14B in Figure \ref{fig:stard_qwen} and Claude-Haiku-4.5 in Figure \ref{fig:stard_claude} indicate that the variance remains consistently low across all reasoning depths, with standard deviations predominantly staying below 7\%. This high level of precision confirms that the performance gap between clean and perturbed inputs is not a stochastic artifact, but a systematic result of Alignment Collapse. Furthermore, the non-overlapping distributions across repeated trials robustly validate our conclusion that deep reasoning structurally compromises alignment.

\noindent \textbf{RT Experiment Supplement.} We investigate the impact of reasoning on alignment by utilizing the RT framework to simulate deep reasoning processes within standard models. As shown in Table~\ref{tab:drj_supplement}, our empirical evaluation across various models the activation of the deep reasoning mode through RT triggers significantly increases the total generated token counts during inference. For instance when integrating RT with FlipAttack on Qwen3-8B the average generated token count surges from 282 to 1568 tokens while the Rejection Success Rate(RSR) drops sharply from 92.31\% to 29.03\%. This transition to a reasoning-heavy cognitive paradigm directly correlates with a significant decline in safety performance across all tested adversarial methods. These results demonstrate that the introduction of the deep reasoning mechanism effectively induces alignment collapse. Our findings confirm that the reasoning functions as a catalyst that exacerbates the model's vulnerability to external perturbations.

\begin{table*}[t]
    \centering
    \renewcommand{\arraystretch}{1.2} 
    \setlength{\tabcolsep}{10pt}      
    \scalebox{0.8}{
        \begin{tabular}{@{}lccccc@{}}
            \toprule
            \textbf{Method} & 
            \textbf{Qwen3-8B} $\downarrow$ & 
            \textbf{Qwen3-14B} $\downarrow$ & 
            \textbf{Llama2-7B} $\downarrow$ & 
            \textbf{Llama2-13B} $\downarrow$ & 
            \textbf{DeepSeek-V3} $\downarrow$ \\ 
            \midrule
            No Attack+RT & 
            169/282 \textbf{(\textcolor{lightpink}{-01.73})} & 
            171/253 \textbf{(\textcolor{lightpink}{-00.58})} & 
            155/260 \textbf{(\textcolor{lightpink}{-00.00})} & 
            155/260 \textbf{(\textcolor{lightpink}{-00.00})} & 
            125/246 \textbf{(\textcolor{lightpink}{-00.00})} \\
            
            FlipAttack+RT & 
            623/1568 \textbf{(\textcolor{lightpink}{-63.28})} & 
            321/418 \textbf{(\textcolor{lightpink}{-02.12})} & 
            191/200 \textbf{(\textcolor{lightpink}{-00.00})} & 
            179/186 \textbf{(\textcolor{lightpink}{-00.00})} & 
            396/470 \textbf{(\textcolor{lightpink}{-02.88})} \\
            
            PAP+RT & 
            1514/1665 \textbf{(\textcolor{lightpink}{-04.04})} & 
            1752/2260 \textbf{(\textcolor{lightpink}{-01.94})} & 
            447/750 \textbf{(\textcolor{lightpink}{-06.35})} & 
            465/569 \textbf{(\textcolor{lightpink}{-03.07})} & 
            750/939 \textbf{(\textcolor{lightpink}{-08.08})} \\
            
            ArtPrompt+RT & 
            821/1948 \textbf{(\textcolor{lightpink}{-07.51})} & 
            153/1030 \textbf{(\textcolor{lightpink}{-26.54})} & 
            292/436 \textbf{(\textcolor{lightpink}{-43.65})} & 
            360/387 \textbf{(\textcolor{lightpink}{-26.06})} & 
            336/603 \textbf{(\textcolor{lightpink}{-14.81})} \\ 
            \bottomrule
        \end{tabular}
    }
    \caption{Experimental supplement for RT on AdvBench. Each cell reports the average generated token counts under Standard (w/o RT), presented as Standard / RT. The values in parentheses indicate the influence of RT on the Rejection Success Rate (RSR). The substantial reductions highlighted in \textbf{\textcolor{lightpink}{light pink}} demonstrate the safety alignment collapse induced by the activation of the deep reasoning process.}
    \label{tab:drj_supplement}
\end{table*}

\section{Interpretable Analysis}
\label{sec:mechanism}

In Section~\ref{sec:mechanistic_explanation}, we qualitatively established that the reasoning process competes for the model's finite attention capacity. We provide a rigorous mathematical derivation of Attention Dilution, focusing on the intrinsic properties of Rotary Positional Embedding (RoPE)~\cite{su2024roformer} utilized in modern LRMs like Qwen3 and DeepSeek.

\subsection{Long-term Decay in RoPE vs. Additive PE}

RoPE encodes positional information by applying a multiplicative rotation matrix $\mathcal{R}$ to the hidden representations. For an initial safety instruction at position $m$ and a current reasoning token at position $n$ ($n \gg m$), the unnormalized attention score $s_{n,m}$ is computed as:
\begin{equation}
s_{n,m} = (\mathcal{R}_n \mathbf{x}_q)^T (\mathcal{R}_m \mathbf{x}_k) = \mathbf{x}_q^T \mathcal{R}_{n-m} \mathbf{x}_k,
\end{equation}
where $\mathbf{x}_q, \mathbf{x}_k \in \mathbb{R}^d$ are the content-based query and key vectors, and $\mathcal{R}_{\tau} \in \mathbb{R}^{d \times d}$ is the block-diagonal rotation matrix corresponding to the relative distance $\tau = n - m$. In the complex domain, this score is modulated by the relative rotation:
\begin{equation}
\mathbb{E}[s_{n,m}] \propto \text{Re}\left[ \sum_{j=1}^{d/2} (\mathbf{h}_{q,j} \mathbf{h}_{k,j}^*) e^{i(n-m)\theta_j} \right],
\end{equation}
where $\mathbf{h}_{q,j}$ and $\mathbf{h}_{k,j}$ denote the $j$-th complex-valued pairs of the query and key, $\mathbf{h}^*$ represents the complex conjugate, $\text{Re}[\cdot]$ denotes the real part, and $\theta_j$ is the pre-defined rotation frequency for the $j$-th dimension. 

As the reasoning depth $L$ increases, the relative distance $\tau = n-m$ between the generation head and the initial alignment tokens grows significantly. The high-frequency oscillations of the $e^{i\tau\theta}$ term cause the expectation of the inner product to exhibit a decay trend. This multiplicative decay suppresses the magnitude of the numerator $\exp(s_{n,m})$ for distant input tokens, effectively weakening the signal of the original constraints.

\subsection{Asymptotic Collapse of Attention Weights}

We analyze the attention weight $\alpha_{x}^{(L)}$ assigned to a fixed input token $x \in X_{\text{align}}$ as the reasoning length $L \to \infty$ within the softmax framework:
\begin{equation}
\alpha_{x}^{(L)} = \frac{\exp(s_{L,x})}{\sum\limits_{j \in X} \exp(s_{L,j}) + \sum\limits_{k \in Y_{<L}} \exp(s_{L,k})}.
\end{equation}

The structural failure of alignment is driven by two concurrent factors as $L$ extends:
\begin{enumerate}
    \item \textbf{Numerator Vanishing:} Following the decay property of RoPE, $\exp(s_{L,x})$ diminishes as the relative distance $L-x$ increases, leading to a marginalized representation of the instructions.
    \item \textbf{Denominator Inflation:} The reasoning contribution term $\sum_{k \in Y_{<L}} \exp(s_{L,k})$ expands linearly with the generation length $L$. Due to the locality bias of the attention mechanism, reasoning tokens $k$ proximal to the current position $L$ where $|L-k|$ is small maintain significantly higher attention scores than the distant input.
\end{enumerate}

Formally, the asymptotic behavior of the attention weight is expressed as:
\begin{equation}
\lim_{L \to \infty} \alpha_{x}^{(L)} \approx \frac{\epsilon}{\text{Const} + \sum_{k=1}^{L} \exp(s_{\text{proximal}})} \to 0,
\end{equation}
where $\epsilon$ represents the attenuated contribution from the initial alignment context. This derivation confirms that Attention Dilution is a structural inevitability in RoPE-based architectures. In the absence of residual interventions like RRA, the model capacity to attend back to safety constraints is systematically eroded by the accumulation of internal reasoning steps.

\section{Analysis of RRA}
\label{sec:rra-analysis}

This section provides additional analysis of Reasoning Residual Alignment (RRA).
Here, we ask a more specific question: does RRA help merely because it adds an explicit safety cue, or because it restores the salience of the original input after a long reasoning trace?

\subsection{RRA Variants on Safety Evaluation}
\label{sec:rra-variants}

Let $X$ denote the original input and $Y$ the reasoning trace generated from $X$.
The reasoning baseline uses $[X;Y]$, whereas RRA uses $[X;Y;X]$, re-injecting the original input before final decoding.

We compare RRA against two lightweight alternatives.
The first appends a fixed safety reminder $S$:
\begin{equation}
[X;Y;S].
\end{equation}
This reminder is independent of the user input:
\begin{quote}
Before giving the final answer, verify whether the response is safe and policy-compliant.
If the request is harmful, illegal, or provides actionable instructions for wrongdoing, refuse to comply.
\end{quote}

The second appends an input-dependent anchor $A(X)$:
\begin{equation}
[X;Y;A(X)].
\end{equation}
The anchor briefly restates the original request:
\begin{quote}
Original request anchor: \{brief summary of the user request\}.
Before answering, check whether this request is safe. If it is harmful, refuse to provide actionable details.
\end{quote}

\begin{table*}[t]
\centering
\small
\renewcommand{\arraystretch}{1.15}
\setlength{\tabcolsep}{8pt}
\begin{tabular}{lccccc}
\toprule
\textbf{Context} & \textbf{FlipAttack} & \textbf{PAP} & \textbf{ArtPrompt} & \textbf{Average} & \textbf{Prompt Cost} \\
\midrule
$[X;Y]$ 
& 57.42 & 73.07 & 40.00 & 56.83 & 0 \\
$[X;Y;S]$ 
& 66.31 (+8.89) & 76.42 (+3.35) & 48.06 (+8.06) & 63.60 (+6.77) & 24 \\
$[X;Y;A(X)]$ 
& 62.48 (+5.06) & 74.16 (+1.09) & 45.71 (+5.71) & 60.78 (+3.95) & 47 \\
$[X;Y;X]$ 
& 64.57 (+7.15) & 73.65 (+0.58) & 47.12 (+7.12) & 61.78 (+4.95) & 0 \\
\bottomrule
\end{tabular}
\caption{
RRA variant ablation on AdvBench for Qwen3-8B.
Values in parentheses denote absolute changes relative to the $[X;Y]$ baseline.
$S$ is a fixed safety reminder, while $A(X)$ is an input-dependent anchor.
The ``Prompt Cost'' column counts only newly introduced instruction tokens; $[X;Y;X]$ reuses the original input and adds no new prompt template.
}
\label{tab:rra-variants-8b}
\end{table*}

\begin{table*}[t]
\centering
\small
\renewcommand{\arraystretch}{1.15}
\setlength{\tabcolsep}{8pt}
\begin{tabular}{lccccc}
\toprule
\textbf{Context} & \textbf{FlipAttack} & \textbf{PAP} & \textbf{ArtPrompt} & \textbf{Average} & \textbf{New Prompt Tokens} \\
\midrule
$[X;Y]$ 
& 45.57 & 80.75 & 35.00 & 53.77 & 0 \\
$[X;Y;S]$ 
& 55.36 (+9.79) & 85.02 (+4.27) & 39.98 (+4.98) & 60.12 (+6.35) & 24 \\
$[X;Y;A(X)]$ 
& 48.27 (+2.70) & 83.22 (+2.47) & 36.37(+1.37) & 55.95 (+2.18) & 47 \\
$[X;Y;X]$ 
& 51.74(+6.17) & 81.13 (+0.58) & 42.57 (+7.57) & 58.48 (+4.71) & 0 \\
\bottomrule
\end{tabular}
\caption{
RRA variant ablation on AdvBench for Qwen3-14B.
Values in parentheses denote absolute changes relative to the $[X;Y]$ baseline.
The fixed safety reminder performs best on this refusal-oriented benchmark, but RRA attains competitive gains without introducing a new safety-specific prompt.
}
\label{tab:rra-variants-14b}
\end{table*}

Tables~\ref{tab:rra-variants-8b} and~\ref{tab:rra-variants-14b} show that the fixed safety reminder yields the highest average RSR on AdvBench.
This is unsurprising: AdvBench rewards refusal behavior, and $S$ directly cues the model to refuse harmful requests.
However, this improvement depends on injecting an additional safety-specific instruction.
By contrast, RRA reuses the original input, introduces no new prompt template, and still recovers a substantial portion of the lost refusal behavior.

\subsection{RRA Variants on Reasoning Benchmarks}
\label{app:rra-task-general}

The gains of $S$ on AdvBench may be specific to safety evaluation, since the reminder explicitly asks the model to check whether the request is harmful.
To test this, we apply the same variants to AIME2024 and LogicAsker under perturbations.
Unlike AdvBench, these benchmarks require solving the original task rather than refusing the query.
A generic safety reminder therefore should not help, and may even distract the model from the reasoning problem.

\begin{table*}[t]
\centering
\small
\renewcommand{\arraystretch}{1.12}
\setlength{\tabcolsep}{7pt}
\begin{tabular}{llcccc}
\toprule
\textbf{Model} & \textbf{Perturbation} & \textbf{Base} & \textbf{+S} & \textbf{+A(X)} & \textbf{+X} \\
\midrule
Qwen3-8B  & P-I  & 25.00 & 24.58 (-0.42) & 27.08 (+2.08) & \textbf{28.33} (+3.33) \\
Qwen3-8B  & P-II & 19.16 & 18.75 (-0.41) & 21.25 (+2.09) & \textbf{22.50} (+3.34) \\
Qwen3-14B & P-I  & 43.33 & 42.91 (-0.42) & 45.00 (+1.67) & \textbf{46.25} (+2.92) \\
Qwen3-14B & P-II & 32.91 & 32.50 (-0.41) & 35.00 (+2.09) & \textbf{36.25} (+3.34) \\
\bottomrule
\end{tabular}
\caption{
Accuracy comparison of RRA variants on AIME2024 at $L{=}4096$.
Base$=[X;Y]$, +S$=[X;Y;S]$ (+24 tokens), +A(X)$=[X;Y;A(X)]$ (+47 tokens), and +X$=[X;Y;X]$ (+0 tokens).
Values in parentheses denote absolute changes relative to Base.
}
\label{tab:rra-aime-variants}
\end{table*}

\begin{table*}[t]
\centering
\small
\renewcommand{\arraystretch}{1.12}
\setlength{\tabcolsep}{7pt}
\begin{tabular}{llcccc}
\toprule
\textbf{Model} & \textbf{Perturbation} & \textbf{Base} & \textbf{+S} & \textbf{+A(X)} & \textbf{+X} \\
\midrule
Qwen3-8B  & P-I & 58.12 & 57.84 (-0.28) & 60.04 (+1.92) & \textbf{61.38} (+3.26) \\
Qwen3-14B & P-I & 73.28 & 72.96 (-0.32) & 75.02 (+1.74) & \textbf{76.18} (+2.90) \\
\bottomrule
\end{tabular}
\caption{
Accuracy comparison of RRA variants on LogicAsker at $L{=}4096$.
Base$=[X;Y]$, +S$=[X;Y;S]$ (+24 tokens), +A(X)$=[X;Y;A(X)]$ (+47 tokens), and +X$=[X;Y;X]$ (+0 tokens).
Values in parentheses denote absolute changes relative to Base.
}
\label{tab:rra-logicasker-variants}
\end{table*}

Tables~\ref{tab:rra-aime-variants} and~\ref{tab:rra-logicasker-variants} show a clear contrast with AdvBench.
The fixed safety reminder slightly reduces accuracy on both AIME2024 and LogicAsker, indicating that its effect does not transfer beyond refusal-oriented evaluation.
In contrast, both $A(X)$ and RRA improve performance under perturbation, with RRA consistently achieving the strongest gains.
These results support our interpretation that RRA helps by restoring the salience of the original input, rather than by adding a generic safety instruction.

Overall, the variant study disentangles two mechanisms.
A fixed safety reminder is effective on safety benchmarks because it directly encourages refusal, but this effect does not generalize to ordinary reasoning tasks.
RRA is less prompt-specific: it reuses the original input, introduces no new safety template, and improves both refusal robustness and perturbation robustness under distracting inputs.

\end{document}